\PassOptionsToPackage{hyphens}{url}
\PassOptionsToPackage{most}{tcolorbox}
\documentclass[11pt]{article}
\let\standardaddcontentsline\addcontentsline
\usepackage[preprint]{arxiv}
\let\addcontentsline\standardaddcontentsline
\usepackage{times}
\usepackage[utf8]{inputenc}
\usepackage[T1]{fontenc}
\usepackage{graphicx}
\usepackage{nicefrac}
\usepackage{microtype}
\usepackage{xspace}

\usepackage{algorithm}
\usepackage{algpseudocode}
\usepackage{amsmath}
\usepackage{amssymb}
\usepackage{bm}
\usepackage{booktabs}
\usepackage{array}
\usepackage{tabularx}
\usepackage{multirow}
\usepackage{colortbl}
\usepackage{bbding}
\usepackage{tikz}
\usepackage{placeins}
\usetikzlibrary{arrows.meta,positioning,fit,calc}
\hypersetup{
hypertexnames=false,
pdftitle={World-to-Wrist: Task-Conditioned Future Wrist Modeling for Fine-Grained Robot Manipulation},
pdfauthor={Yuhao Pan, Haosong Peng, Zhengshen Zhang, Zhengyang Yan, Yalun Dai, Fushuo Huo, Chujie Wang, Tianyu Qi, Xiucheng Wang, Nan Cheng, Wenchao Xu}
}

\DeclareMathOperator*{\E}{\mathbb{E}}
\newcommand{\ours}{\mbox{\(\mathsf{W}^{2}\)-VLA}\xspace}
\newcommand{\CoT}{\mbox{\(\mathsf{W}^{2}\)-CoT}\xspace}
\newcommand{\stopgrad}{\operatorname{sg}}

\newcommand{\projectpageurl}{https://xxx}
\newcommand{\correspondingemail}{wenchaoxu@ust.hk}
\definecolor{m2wblue}{RGB}{55,103,177}
\definecolor{m2wgreen}{RGB}{69,145,109}
\definecolor{m2wamber}{RGB}{213,143,52}
\definecolor{m2wred}{RGB}{190,80,70}
\definecolor{m2wgray}{RGB}{96,104,116}
\definecolor{linecolor1}{RGB}{199,216,238}
\definecolor{linecolor2}{RGB}{196,224,207}
\newcommand{\liberocaptionbox}[2]{\begingroup\setlength{\fboxsep}{0.75pt}\raisebox{0pt}[0pt][0pt]{\colorbox{#1}{#2}}\endgroup}

\title{World-to-Wrist: Task-Conditioned Future Wrist Modeling for Fine-Grained Robot Manipulation}

\author{
\normalfont
Yuhao Pan\textsuperscript{1,*}
\quad
Haosong Peng\textsuperscript{1,*}
\quad
Zhengshen Zhang\textsuperscript{2}
\quad
Zhengyang Yan\textsuperscript{1}
\\
\normalfont
Yalun Dai\textsuperscript{3}
\quad
Fushuo Huo\textsuperscript{7}
\quad
Chujie Wang\textsuperscript{4}
\quad
Tianyu Qi\textsuperscript{5}
\\
\normalfont
Xiucheng Wang\textsuperscript{6}
\quad
Nan Cheng\textsuperscript{6}
\quad
Wenchao Xu\textsuperscript{1,\ensuremath{\dagger}}
\\[3pt]
\normalfont\small
\textsuperscript{1}The Hong Kong University of Science and Technology
\quad
\textsuperscript{2}National University of Singapore
\\
\normalfont\small
\textsuperscript{3}Nanyang Technological University
\quad
\textsuperscript{4}Wuhan University
\quad
\textsuperscript{5}Sun Yat-sen University
\\
\normalfont\small
\textsuperscript{6}Xidian University
\quad
\textsuperscript{7}Southeast University
\\[2pt]
\normalfont\small
\textsuperscript{*}\,Equal contribution.\quad
\textsuperscript{\ensuremath{\dagger}}\,Corresponding author.
}

\begin{document}

\setlength{\headheight}{24pt}
\fancypagestyle{firstpage}{
  \fancyhf{}
  \lhead{%
    \includegraphics[height=18pt]{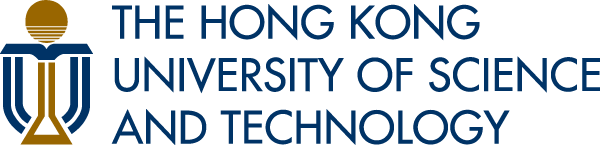}%
    \hspace{0.7em}%
    \includegraphics[height=18pt]{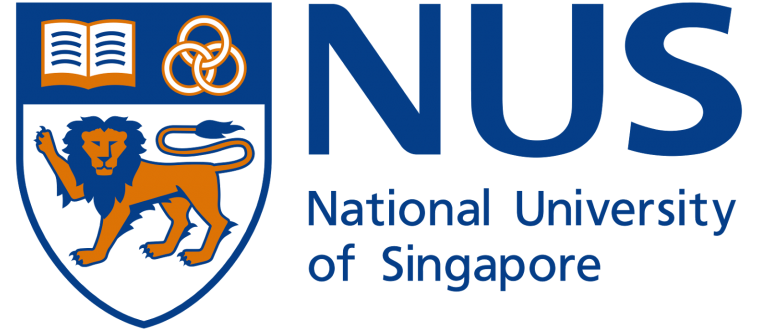}%
    \hspace{0.7em}%
    \includegraphics[height=18pt]{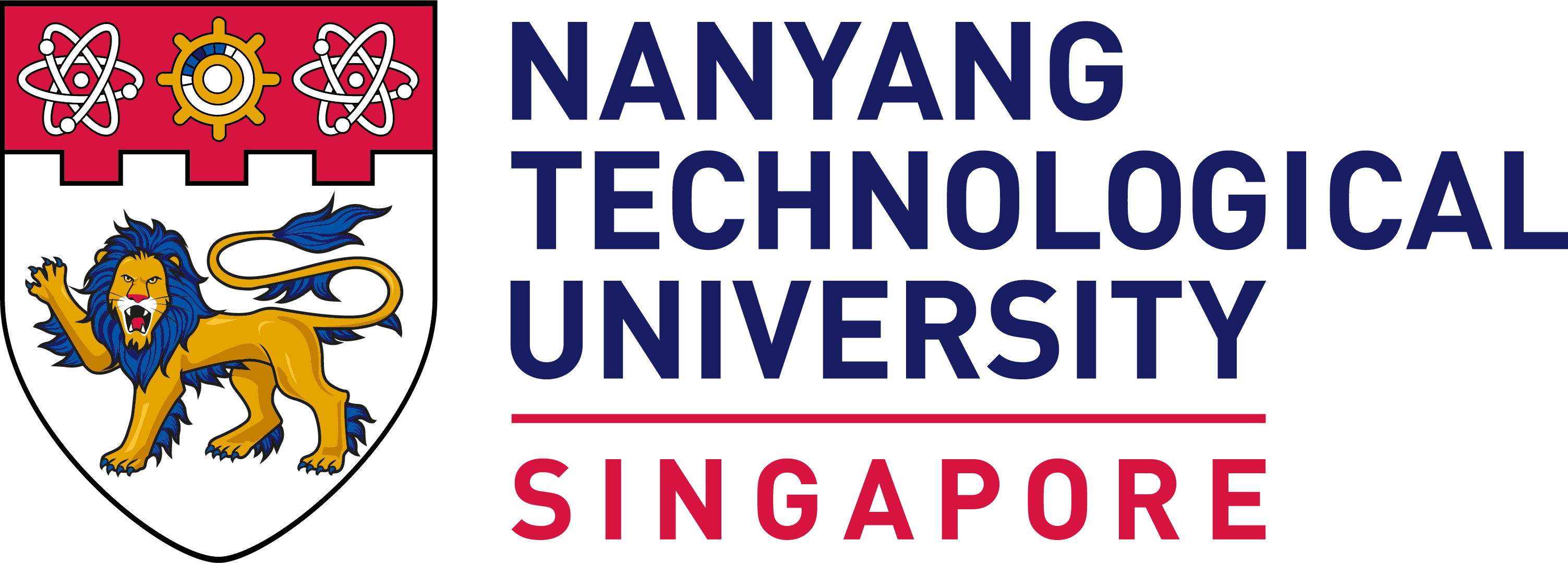}%
  }
  \rhead{\small\textcolor[HTML]{4D4D4D}{\today\quad Preprint}}
  \cfoot{\thepage}
  \renewcommand{\headrulewidth}{0.4pt}
}

\maketitle
\vspace{-5mm}
\begin{abstract}

Vision-language-action (VLA) models often treat main-view and wrist-view observations as parallel visual inputs, overlooking their distinct roles in robot manipulation.
Fine-grained manipulation, however, benefits from anticipating how wrist-local interactions may evolve under the global task context.
To address this limitation, we present \textbf{World-to-Wrist VLA (\ours)}, a VLA model for fine-grained robot manipulation with task-conditioned future wrist modeling.
Given current multi-view observations and an instruction, \ours contextualizes a set of latent modeling tokens as a compact interface between the VLM and the wrist predictor.
Conditioned on this interface and wrist history, the wrist predictor forecasts future wrist latents, which are converted into future-aware context for action prediction.
In addition, we propose \CoT, a synthesis pipeline that produces structured annotations for manipulation progress, physical transition cues, and wrist-local evidence.
These structured annotations provide auxiliary supervision to help shape the
task-conditioned latent interface.
Experiments on LIBERO, LIBERO-Plus, RoboTwin 2.0, and real-world manipulation tasks demonstrate
improved fine-grained and contact-sensitive manipulation across single-arm and bimanual
settings, while maintaining real-time action generation above $80$~Hz.

\par\medskip
\noindent{\small
\begin{tabular}{@{}l@{\qquad}l@{}}
\textbf{Correspondence:} &
\href{mailto:\correspondingemail}{\texttt{\correspondingemail}} \\
\textbf{Project Page:} &
\href{\projectpageurl}{\url{https://yyyyu120.github.io/W2-VLA/}}
\end{tabular}
}

\end{abstract}

\addtocontents{toc}{\protect\setcounter{tocdepth}{-1}}
\section{Introduction}
\begin{figure}[!t]
\centering
\includegraphics[width=\linewidth]{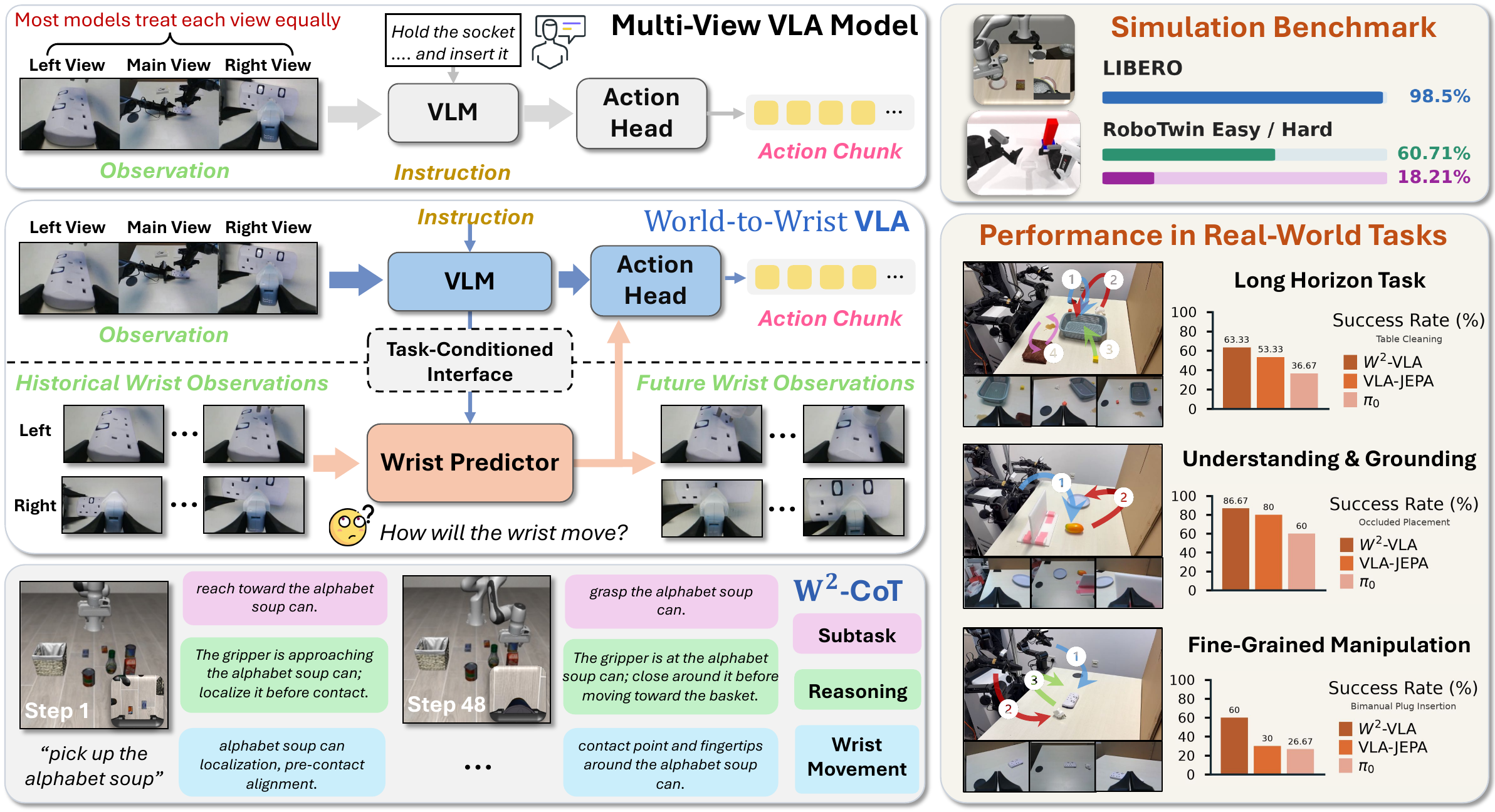}
\caption{We propose \textbf{\ours}, a vision-language-action model for fine-grained robot manipulation based on task-conditioned future wrist modeling.
Conventional methods generate actions conditioned on current multi-view observations and a language instruction, whereas \ours additionally predicts future wrist latents by combining a task-conditioned interface with wrist history.
}
\label{fig:teaser}
\end{figure}

Vision-language-action (VLA) models provide a unified interface for mapping visual observations and language instructions to robot actions~\cite{brohan2022rt1,zitkovich2023rt,kim2024openvla,octo2024,black2024pi_0}.
Recent advances in large-scale pretraining, cross-embodiment transfer, and continuous action generation have substantially improved their generality and control capabilities~\cite{intelligence2025pi_,bjorck2025gr00t,wen2025dexvla,zheng2025x,wang2026vla}.
However, as shown in Fig.~\ref{fig:teaser}, existing multi-view VLA models typically encode or fuse different camera views as parallel visual inputs~\cite{shukor2025smolvla,wen2025dexvla,kim2025fine}, without explicitly modeling the action-proximal role of wrist observations.
Specifically, while the main view provides global task context through scene layout, object identity, goal relations, and task progress, wrist observations directly reveal rapidly changing gripper--object interactions around the end effector.
This distinction is especially critical for fine-grained manipulation tasks, such as plug insertion, which require close coordination between global task understanding and local interaction dynamics.
This action-proximal role motivates treating wrist observations as more than
another current visual input.

A natural way to exploit this wrist-specific role is to model not only the current wrist observation, but also its near-term evolution.
Recent future-predictive visuomotor models obtain such foresight by forecasting visual subgoals, future observations, or latent representations~\cite{zhao2025cot,shouhalo,sun2026vla,
zhang2026dreamvla,luo2026being}.
However, these targets are often defined over the global scene or unified policy representations, without isolating temporal changes around the end effector.
Wrist-centered latent prediction instead provides an action-proximal target that captures how local interactions may evolve beyond the current state, while focusing on end-effector-relevant changes without reconstructing pixel-level appearance.

However, wrist-centered prediction remains task-ambiguous: similar wrist histories may correspond to multiple plausible futures, and global task context is needed to identify the relevant one.
For example, an approach may lead to grasping or pushing, while an aligned configuration may call for continued adjustment, insertion, or release, depending on the instruction, global scene context, and current manipulation stage.
Wrist history therefore provides evidence about the current local interaction, whereas global task context is needed to identify the task-relevant transition.
This motivates a compact task-conditioned interface, shaped by structured semantic supervision, that connects VLM-encoded global task context to future wrist prediction.

We therefore present \textbf{World-to-Wrist VLA (\ours)}, a VLA model that formulates future wrist modeling as task-conditioned latent prediction.
Given the current multi-view observations and an instruction, the VLM contextualizes a dedicated set of latent modeling tokens.
The hidden states of these tokens form a fixed-length, task-conditioned interface between the VLM and the wrist predictor.
To shape task-conditioned VLM representations, we construct structured \textbf{$\mathsf{W}^{2}$-Chain-of-Thought (\CoT)} annotations through a synthesis pipeline, providing supervision for \textit{manipulation progress}, \textit{physical transition cues}, and \textit{wrist-local evidence}.
During training, this interface is shaped by autoregressive annotation prediction, future wrist prediction, and action generation.
Conditioned on this interface and wrist history encoded by a frozen V-JEPA~2.1 encoder~\cite{mur2026v}, the predictor forecasts task-relevant future wrist latents.
A lightweight adapter extracts future-aware wrist context from the predicted latents and combines it with the VLM context to condition a flow-matching action head~\cite{peebles2023scalable}.
Inference requires neither future wrist observations nor autoregressive \CoT decoding.

We evaluate \ours on LIBERO, LIBERO-Plus, RoboTwin~2.0, and three real-world tasks spanning single-arm and bimanual manipulation.
\ours achieves $98.5\%$ average success on LIBERO and $84.0\%$ zero-shot success on LIBERO-Plus.
It further reaches $60.71\%$ and $18.21\%$ under the RoboTwin~2.0 Easy and Hard settings, respectively, while outperforming all baselines across the three real-world tasks under standard and out-of-distribution (OOD) conditions.
The main contributions are summarized as follows:

\begin{enumerate}
    \item We formulate future wrist modeling as task-conditioned latent
    prediction, defining a World-to-Wrist pathway from global task context
    to future wrist-local dynamics for fine-grained control.

    \item We develop an evidence-grounded \CoT synthesis pipeline whose
    structured annotations help shape a fixed-length task-conditioned
    interface that guides future wrist latent prediction from wrist history.
    The policy does not require CoT generation at inference time, enabling real-time action generation at over 80 Hz.

    \item We evaluate \ours on LIBERO, LIBERO-Plus, RoboTwin~2.0, and three real-world manipulation tasks spanning long-horizon execution and fine-grained manipulation.
    \ours achieves $84.0\%$ zero-shot success on LIBERO-Plus and demonstrates strong performance across single-arm, bimanual, standard, and OOD settings.
\end{enumerate}
\section{Related Work}

\noindent\textbf{Generalist VLA models.}
Vision-language-action models learn unified mappings from visual observations and language instructions to robot actions.
RT-1~\cite{brohan2022rt1} and RT-2~\cite{zitkovich2023rt} established scalable transformer policies for real-world manipulation and demonstrated the transfer of web-scale knowledge to robot control.
OpenVLA~\cite{kim2024openvla} and Octo~\cite{octo2024} improved the accessibility and adaptability of generalist robot policies.
Recent systems further advance continuous action generation through flow-matching policies~\cite{black2024pi_0,intelligence2025pi_}, diffusion-based action experts~\cite{wen2025dexvla}, and foundation models for humanoid and cross-embodiment control~\cite{bjorck2025gr00t,zheng2025x}.
StarVLA~\cite{community2026starvla} provides modular interfaces between VLM backbones and action heads.
Complementary to these advances, \ours introduces a task-conditioned interface that connects the current VLM context to future wrist prediction.

\noindent\textbf{Cross-view modeling and future-aware prediction.}
Cross-view policies combine external and wrist observations through cross-view attention~\cite{jangir2022look} or adaptive view weighting~\cite{lan2025bfa}.
In a complementary direction, WristWorld~\cite{qian2025wristworld} explores wrist-view video generation from anchor views.
Related spatial foundation models incorporate auxiliary depth and camera parameters to enrich multi-view geometric representations for VLA models~\cite{peng2025omnivggt}.
Future-aware objectives provide predictive supervision for task-relevant state evolution.
DreamVLA~\cite{zhang2026dreamvla} forecasts dynamic, spatial, and semantic world knowledge, while VLA-JEPA~\cite{sun2026vla} learns action-relevant representations through future latent prediction.
WoG~\cite{su2026world} converts future observations into compact conditioning representations for action generation, and Being-H0.7~\cite{luo2026being} aligns states inferred from context with future-informed latent representations.
Collectively, these works advance cross-view fusion, wrist-view generation, and future-aware prediction.
However, these works generally do not explicitly isolate wrist-local temporal dynamics as a dedicated prediction target jointly conditioned on global task context and wrist history.
\ours addresses this gap by formulating future wrist modeling as task-conditioned latent prediction through a fixed-length interface between the VLM and wrist predictor.

\begin{figure}[!t]
\centering
\includegraphics[width=\textwidth]{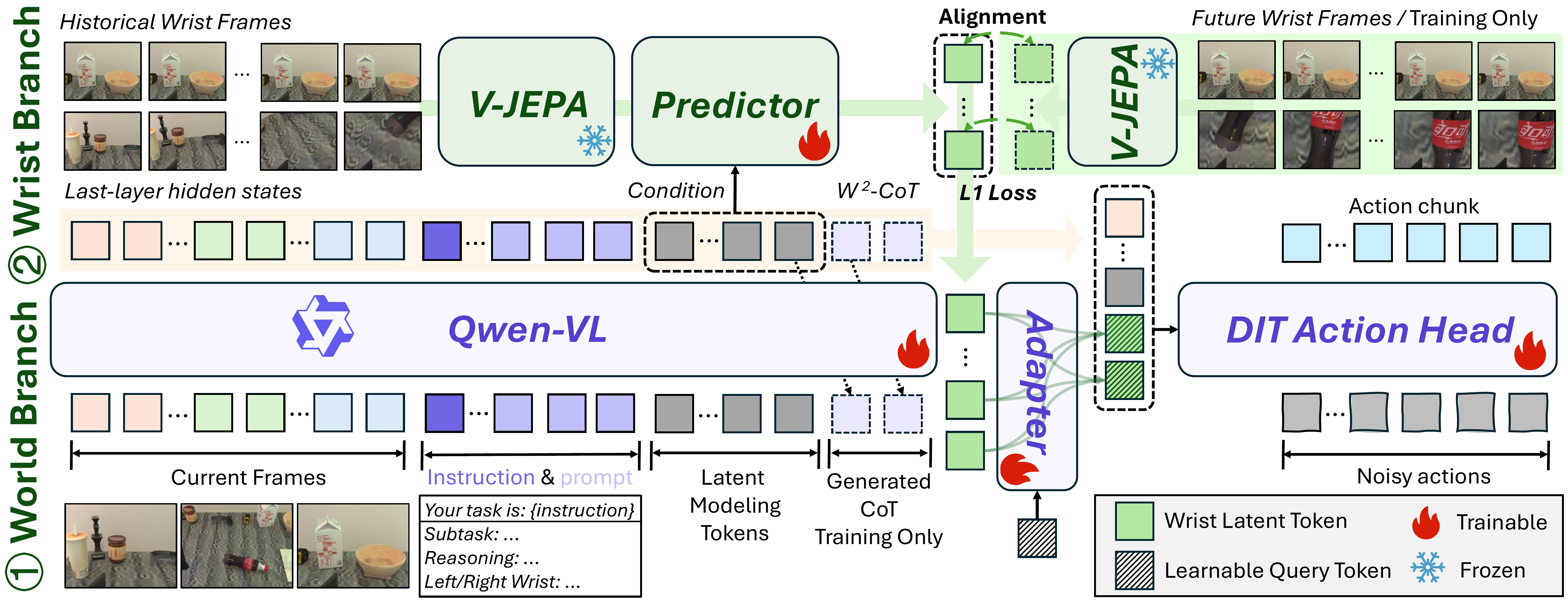}
\caption{\textbf{Overview of \ours.}
\textcircled{1}~The VLM contextualizes latent modeling tokens using current multi-view observations and an instruction, forming a fixed-length task-conditioned interface jointly shaped by \CoT annotation supervision, future wrist prediction, and action generation objectives.
\textcircled{2}~Conditioned on this interface and wrist-history features encoded by V-JEPA~2.1, a predictor forecasts future wrist latents, which a lightweight adapter converts into future-aware wrist context for flow-matching action generation.
Future wrist clips are encoded only to form training targets, and inference requires no \CoT decoding.}
\label{fig:overview}
\end{figure}

\noindent\textbf{Intermediate supervision and structured grounding.}
Intermediate targets provide VLA models with structured learning signals beyond action labels.
CoT-VLA~\cite{zhao2025cot} predicts future visual goals before actions, while HALO~\cite{shouhalo} combines textual task reasoning, visual foresight, and action prediction.
GraphCoT-VLA~\cite{huang2026graphcot} introduces structured spatial reasoning, and related grounding methods incorporate spatial representations, priors, traces, and hierarchical structures~\cite{qu2025spatialvla,zhang2025spatial,spatialforcing2025,zheng2025tracevla,yang2026hivla}.
LaRA-VLA~\cite{bai2026latent} represents intermediate reasoning in continuous latent states to avoid explicit decoding.
RoboInter~\cite{li2026robointer} provides dense frame-level intermediate annotations that connect manipulation planning and execution.
Collectively, these methods demonstrate that linguistic, visual, spatial, and latent intermediate structures can supplement action supervision.
Building on this line of work, \ours uses structured \CoT annotations as auxiliary training targets to help shape a fixed-length task-conditioned interface for future wrist prediction, without requiring annotation decoding at inference.

\section{Methodology}

Fig.~\ref{fig:overview} illustrates the overall architecture of \ours.
At each control step, the VLM contextualizes a dedicated set of latent modeling tokens using the current multi-view observations and instruction, forming a fixed-length task-conditioned interface.
Together with encoded wrist history, this interface conditions the prediction of future wrist latents, which are converted into future-aware wrist context for action generation.
In the following subsections, we detail the latent modeling interface, task-conditioned future wrist prediction, structured \CoT auxiliary supervision, and the joint training objectives alongside the inference procedure.

\subsection{Task-Conditioned Latent Modeling Interface}
\label{sec:latent_interface}
We instantiate the task-conditioned interface using $K$ dedicated latent
modeling tokens contextualized by the current multi-view observations and
instruction, where $K$ is a hyperparameter.
Specifically, at control step $t$, let $\ell$ denote the instruction,
$\mathbf{I}_t^m$ the current main-view observations, and $\mathbf{I}_t^w$ the
available wrist-view observations.
We denote the complete visual input by
$\mathcal{O}_t=\{\mathbf{I}_t^m,\mathbf{I}^w_t\}$.
Let $\langle q_k\rangle$ denote the $k$-th latent modeling token and
$\mathcal{P}_q$ the set of corresponding token positions.
We append the latent modeling token sequence to the instruction prompt:
\begin{align}
    \mathbf{p}_t
    &=
    \left[
    \operatorname{Prompt}(\ell)
    \mid
    \langle q_1\rangle
    \mid
    \cdots
    \mid
    \langle q_K\rangle
    \right],
    \label{eq:modeling_prompt}
    \\
    \mathbf{S}_t
    &=
    F_{\theta}^{\mathrm{VLM}}
    \left(
    \mathcal{O}_t,
    \mathbf{p}_t
    \right)
    [\mathcal{P}_q]
    \in \mathbb{R}^{K\times d},
    \label{eq:latent_interface}
\end{align}
where $d$ is the VLM hidden dimension and $\mathbf{S}_t$ denotes the
observation- and instruction-conditioned final-layer hidden states selected
at the latent modeling token positions $\mathcal{P}_q$.
The resulting states serve as a fixed-length task-conditioned interface for
the wrist predictor.
During training, $\mathbf{S}_t$ receives learning signals from auxiliary
annotation supervision, future wrist prediction, and action generation.
These objectives affect the interface through different computational paths,
as described in the following subsections.

\subsection{Task-Conditioned Future Wrist Prediction}
\label{sec:future_wrist}
The wrist branch predicts task-relevant future wrist dynamics from encoded
wrist history, conditioned on the interface states $\mathbf{S}_t$.
Specifically, let $\mathbf{I}^{w}_{t,\mathrm{hist}}$ denote a wrist-history clip ending at
control step $t$, and let $\mathbf{I}^{w}_{t,\mathrm{fut}}$ denote the
corresponding future target clip used only during training.
A frozen V-JEPA~2.1 encoder $E_{\phi}$ maps these clips to historical wrist
tokens and future target tokens:
$\mathbf{Z}^{w}_{t,\mathrm{hist}}
=E_{\phi}(\mathbf{I}^{w}_{t,\mathrm{hist}})$ and
$\mathbf{Z}^{w}_{t,\mathrm{fut}}
=E_{\phi}(\mathbf{I}^{w}_{t,\mathrm{fut}})$.
For multiple wrist views, per-view wrist tokens are concatenated across views
within each latent time step before future prediction.

The predictor repeats $\mathbf{S}_t$ across the V-JEPA latent time steps and
independently maps the interface states and historical wrist tokens to a
shared hidden space.
At each step, the interface tokens are prepended to the corresponding wrist
latent tokens and processed by a bidirectional Transformer:
\begin{align}
    \widehat{\mathbf{Z}}^{w}_{t,\mathrm{fut}}
    =
    G_{\psi}\!\left(
    \mathbf{Z}^{w}_{t,\mathrm{hist}},
    \mathbf{S}_t
    \right).
    \label{eq:future_wrist_prediction}
\end{align}
The encoded future clip provides a detached latent target:
\begin{align}
    \mathcal{L}_{\mathrm{wrist}}
    =
    \left\|
    \widehat{\mathbf{Z}}^{w}_{t,\mathrm{fut}}
    -
    \stopgrad\!\left(
    \mathbf{Z}^{w}_{t,\mathrm{fut}}
    \right)
    \right\|_1.
    \label{eq:wrist_loss}
\end{align}
Here, $\stopgrad(\cdot)$ denotes the stop-gradient operator.
This objective thus only supervises the predictor and provides a learning signal to the
latent modeling interface through its conditioning path.
The predictor therefore models wrist-view state transitions rather than
reconstructing RGB observations.

To aggregate the dense predicted future wrist latents $\mathbf{Z}^{w}_{t,\mathrm{fut}}$ into a fixed number of context tokens, we employ a Q-Former-style~\cite{li2023blip} future-wrist context adapter $A_{\omega}$ with $M$ learnable queries $\mathbf{Q}^{w}=[\mathbf{q}^{w}_1,\ldots,\mathbf{q}^{w}_M]$:
\begin{align}
    \mathbf{C}_t^w
    =
    A_{\omega}\!\left(
    \mathbf{Q}^{w},
    \stopgrad (
    \widehat{\mathbf{Z}}^{w}_{t,\mathrm{fut}})
    \right).
    \label{eq:wrist_context_adapter}
\end{align}
Through cross-attention, the adapter extracts a compact future-aware wrist
context and projects it to the VLM hidden dimension.
The stop-gradient operator prevents the action objective from updating the future wrist predictor through the adapter path, while the adapter itself remains trainable.

\subsection{Structured \texorpdfstring{\CoT}{W2-CoT} Annotation Synthesis and Auxiliary Supervision}
\label{sec:cot_supervision}
We construct structured \CoT annotations and use them as auxiliary training targets.
Each annotation contains three fields.
The \texttt{Subtask} field describes the current manipulation stage and its progress within the task.
The \texttt{Reasoning} field summarizes the robot-centric physical transition supported by the available state--action and visual evidence, such as approach-to-contact, grasp stabilization, transport, alignment, or release.
The \texttt{Wrist} field records wrist-local evidence, including target
proximity, fingertip contact, grasp stability, object motion, alignment,
placement stability, and gripper--object separation.
For bimanual data, \texttt{Wrist} follows the ordered format
\texttt{left=...; right=...}, with each side restricted to evidence about its gripper, contact state, and local motion.


To construct \CoT annotations, we first identify candidate manipulation segments along each trajectory.
Gripper openness, end-effector motion, and action changes indicate boundaries between phases such as approach, grasp, transport, and release.
Visual keyframes sampled around candidate boundaries provide complementary evidence about object identity, contact, placement, and local motion.
For each segment, an offline VLM annotator generates a structured proposal from the task instruction, synchronized main- and wrist-view frames, and state--action evidence.
We then check the proposal for gripper-state consistency, valid temporal ordering, release preconditions, and wrist locality.
For example, an open gripper cannot be labeled as grasping or carrying, and a release stage requires a preceding holding stage.
After verification, we normalize the labels, suppress spurious stages caused by brief jitter or retries, and propagate the segment-level annotations to individual frames.
Only the three normalized fields are rendered as the auxiliary target sequence; internal fields such as phase, target, contact state, confidence, and supporting evidence remain metadata.
Further details are provided in Appendix~\ref{app:cot_construction}.

During training, we optimize an auxiliary next-token prediction objective over the structured annotation sequence
$\mathbf{y}_t^\star
=(y_{t,1}^\star,\ldots,y_{t,N_t}^\star)$:
\begin{align}
    \mathcal{L}_{\mathrm{cot}}
    =
    -\frac{1}{N_t}
    \sum_{n=1}^{N_t}
    \log p_{\theta}\!\left(
        y_{t,n}^\star
        \mid
        \mathcal{O}_t,
        \mathbf{p}_t,
        \mathbf{y}_{t,<n}^\star
    \right).
    \label{eq:cot_loss}
\end{align}
The current observations and instruction jointly contextualize $\mathbf{S}_t$, which serves as a task-conditioned interface for future wrist prediction.
This auxiliary objective further encourages $\mathbf{S}_t$ to capture \textit{manipulation progress}, \textit{physical transition cues}, and \textit{wrist-local evidence}.

\subsection{Training Objective and Inference}
\label{sec:action_training}

Let $\mathcal{P}_{\mathrm{act}}$ denote the visual, instruction, and latent
modeling token positions used for action conditioning, with
$\mathcal{P}_q\subseteq\mathcal{P}_{\mathrm{act}}$.
The resulting VLM action-conditioning context is
\begin{align}
    \mathbf{H}_t^{\mathrm{act}}
    =
    F_{\theta}^{\mathrm{VLM}}
    \left(
        \mathcal{O}_t,
        \mathbf{p}_t
    \right)
    [\mathcal{P}_{\mathrm{act}}].
    \label{eq:act_vlm_context}
\end{align}
It contains visual, instruction, and latent modeling states, while auxiliary
annotation states are excluded.
The same state selection is used during training and inference.
Within $\mathbf{H}_t^{\mathrm{act}}$, the interface states $\mathbf{S}_t$
form the latent modeling subset and also condition future wrist prediction,
while $\mathbf{H}_t^{\mathrm{act}}$ as a whole serves as the VLM component of
the complete action-conditioning context.

We concatenate the VLM context with the future-aware wrist context: $\mathbf{C}_t = \left[ \mathbf{H}_t^{\mathrm{act}} \mid \mathbf{C}_t^w \right]$.
The fused context conditions a DiT-based flow-matching action head, yielding
$\widehat{\mathbf{A}}_t=\Pi_{\eta}(\mathbf{C}_t)$, where $\Pi_{\eta}$ denotes
action generation by integrating the conditional velocity field
$D_{\eta}$.

To train this action head, let $\mathbf{A}_t^\star=[a_t^\star,\ldots,a_{t+H_a-1}^\star]$ denote the target action chunk.
Given Gaussian noise $\bm{\epsilon}\sim\mathcal{N}(\mathbf{0},\mathbf{I})$ and a flow time $\tau\in[0,1]$ sampled from a transformed Beta distribution, we construct the interpolated action state $\mathbf{X}_{\tau}=(1-\tau)\bm{\epsilon}+\tau\mathbf{A}_t^\star$ with target velocity $\mathbf{V}_t^\star=\mathbf{A}_t^\star-\bm{\epsilon}$.
The DiT network estimates the conditional velocity using the flow-matching objective:
\begin{align}
    \mathcal{L}_{\mathrm{act}}
    =
    \E_{\tau,\bm{\epsilon}}
    \left[
        \left\|
        D_{\eta}(
            \mathbf{X}_{\tau},
            \tau,
            \mathbf{C}_t
        )
        -
        \mathbf{V}_t^\star
        \right\|_2^2
    \right].
    \label{eq:action_loss}
\end{align}

The joint objective combines action generation, CoT prediction, and future wrist latent prediction:
\begin{align}
    \mathcal{L}
    =
    \mathcal{L}_{\mathrm{act}}
    +
    \lambda_{\mathrm{cot}}\mathcal{L}_{\mathrm{cot}}
    +
    \lambda_{\mathrm{wrist}}\mathcal{L}_{\mathrm{wrist}},
    \label{eq:total_loss}
\end{align}
where $\lambda_{\mathrm{cot}}$ and $\lambda_{\mathrm{wrist}}$ are
loss-weighting hyperparameters.

At inference, the model receives the current observations $\mathcal{O}_t$, wrist history $\mathbf{I}^{w}_{t,\mathrm{hist}}$, and instruction $\ell$.
The VLM constructs $\mathbf{H}_t^{\mathrm{act}}$ and its latent modeling subset $\mathbf{S}_t$ in a single forward pass.
Conditioned on $\mathbf{S}_t$ and the encoded wrist history, the wrist predictor forecasts future wrist latents, which the adapter converts into the future-aware wrist context $\mathbf{C}_t^w$.
The fused context $\mathbf{C}_t$ then conditions $\Pi_{\eta}$ to generate the action chunk $\widehat{\mathbf{A}}_t$.
Inference requires neither future wrist observations nor autoregressive \CoT decoding.

\section{Experiments}
We evaluate \ours on LIBERO, LIBERO-Plus, RoboTwin~2.0,
and three real-world manipulation tasks on the CoBoT Magic platform.
The experiments assess simulation performance, real-world robustness, component contributions, inference efficiency, and OOD generalization ability.

\subsection{Experimental Setup}
\noindent\textbf{Implementation details.}
We implement \ours based on StarVLA~\cite{community2026starvla}, using
Qwen3-VL-4B-Instruct~\cite{qwen3vl2025} as the vision-language backbone and a DiT-based flow-matching action head.
The action chunk length is 8 for LIBERO and 16 for both RoboTwin~2.0 and
the real-world tasks.
The wrist branch comprises a frozen V-JEPA~2.1 ViT-L/384 encoder, a
four-layer latent predictor, and a lightweight context adapter that maps latent predictions into 32 action-context tokens.
We use 16 latent modeling tokens and structured
\texttt{Subtask}/\texttt{Reasoning}/\texttt{Wrist} targets for auxiliary language modeling during training.
The complete model contains approximately $4.97$B parameters.
Further details on the model implementation and training procedure are
provided in Appendix~\ref{app:implementation-details}.

\begin{table}[t]
\centering

\newcommand{\liberobest}[1]{\liberocaptionbox{linecolor1}{\textbf{#1}}}
\newcommand{\liberosecond}[1]{\liberocaptionbox{linecolor2}{\underline{#1}}}

\caption[Comparison on the LIBERO benchmark]{\textbf{Comparison on the LIBERO benchmark.}
We report the task success rate (\%) for each suite and the average across
all tasks.
\protect\liberocaptionbox{linecolor1}{\textbf{Bold}} denotes the best performance, and
\protect\liberocaptionbox{linecolor2}{\underline{underline}} denotes the second best.
}
\label{table:libero_comparison}

{
\small
\setlength{\tabcolsep}{0pt}
\renewcommand{\arraystretch}{1.12}

\begin{tabular}{
    @{}
    l
    @{\hspace{0.5pt}}c
    @{\hspace{2.35pt}}c
    @{\hspace{2.35pt}}c
    @{\hspace{2.35pt}}c
    @{\hspace{2.35pt}}c
    @{}
}
\toprule
\textbf{Method}
& \textbf{Spatial}
& \textbf{Object}
& \textbf{Goal}
& \textbf{Long}
& \textbf{Avg.} \\
\midrule

OpenVLA~\cite{kim2024openvla}
& 84.7
& 88.4
& 79.2
& 53.7
& 76.5 \\

OpenVLA-OFT~(Kim et al.~\citeyear{kim2025fine})
& 97.6
& 98.4
& 97.9
& 94.5
& 97.1 \\

$\pi_0$~\cite{black2024pi_0}
& 96.8
& 98.8
& 95.8
& 85.2
& 94.2 \\

$\pi_0\texttt{-Fast}$~\cite{pertsch2025fast}
& 96.4
& 96.8
& 88.6
& 60.2
& 85.5 \\

$\pi_{0.5}$~\cite{intelligence2025pi_}
& \liberosecond{98.8}
& 98.2
& \liberosecond{98.0}
& 92.4
& 96.9 \\

Fast-ThinkAct~\cite{huang2026fast}
& 92.0
& 97.2
& 90.2
& 79.4
& 89.7 \\

DreamVLA~\cite{zhang2026dreamvla}
& 97.5
& 94.0
& 89.5
& 89.5
& 92.6 \\

GR00T-N1~(NVIDIA,~\citeyear{bjorck2025gr00t})
& 94.4
& 97.6
& 93.0
& 90.6
& 93.9 \\

MemoryVLA~\cite{shi2025memoryvla}
& 98.4
& 98.4
& 96.4
& 93.4
& 96.7 \\

VLA-JEPA~\cite{sun2026vla}
& 96.2
& \liberosecond{99.6}
& 97.2
& \liberosecond{95.8}
& \liberosecond{97.2} \\

DeepThinkVLA~\cite{yin2025deepthinkvla}
& 96.6
& 99.0
& 96.4
& \liberobest{96.2}
& 97.0 \\

StarVLA~\cite{community2026starvla}
& 97.8
& 98.8
& 97.4
& 92.0
& 96.5 \\

\midrule

\ours
& \liberobest{99.6}
& \liberobest{99.8}
& \liberobest{99.2}
& 95.2
& \liberobest{98.5} \\

\bottomrule
\end{tabular}
}
\end{table}

\begin{table}[t]
\centering

\newcommand{\liberoplusbest}[1]{\liberocaptionbox{linecolor1}{\textbf{#1}}}
\newcommand{\liberoplussecond}[1]{\liberocaptionbox{linecolor2}{\underline{#1}}}

\caption[Zero-shot transfer on LIBERO-Plus]{\textbf{Zero-shot transfer on LIBERO-Plus.}
We report success rates (\%) under seven perturbation types and the overall
success rate across all 10,030 evaluation instances.
Prior results follow the corresponding publications and the LIBERO-Plus
benchmark report~\cite{fei2026liberoplus}.
\protect\liberocaptionbox{linecolor1}{\textbf{Bold}} denotes the best performance, and
\protect\liberocaptionbox{linecolor2}{\underline{underline}} denotes the second best.
}
\label{table:libero_plus}

{
\scriptsize
\setlength{\tabcolsep}{2.8pt}
\renewcommand{\arraystretch}{1.12}

\resizebox{\linewidth}{!}{%
\begin{tabular}{@{}l|cccccccc@{}}
\toprule
\textbf{Method}
& \textbf{Camera}
& \textbf{Robot}
& \textbf{Language}
& \textbf{Light}
& \textbf{Background}
& \textbf{Noise}
& \textbf{Layout}
& \textbf{Overall} \\
\midrule

Mantis~\cite{yang2026mantis}
& 15.7 & 41.8 & 45.9 & 45.1 & 28.9 & 39.2 & 62.5 & 39.8 \\

Fast-WAM~\cite{yuan2026fastwam}
& 16.4 & 44.5 & 68.9 & 78.2 & 53.7 & 37.7 & 60.7 & 51.5 \\

$\pi_0$~\cite{black2024pi_0}
& 13.8 & 6.0 & 58.8 & 85.0 & 81.4 & 79.0 & 68.9 & 53.6 \\

VLA-JEPA~\cite{sun2026vla}
& 63.3 & 67.1 & 85.4 & 95.6 & \liberoplussecond{93.6} & 66.3 & 85.1 & 79.5 \\

OpenVLA-OFT~\cite{kim2025fine}
& 56.4 & 31.9 & 79.5 & 88.7 & 93.3 & 75.8 & 74.2 & 69.6 \\

StarVLA~\cite{community2026starvla}
& 52.5 & 49.8 & 88.5 & 95.7 & \liberoplusbest{95.7} & 73.0 & 76.9 & 74.1 \\

Being-H0.7~\cite{luo2026being}
& \liberoplusbest{82.0} & 59.0 & 82.8 & \liberoplussecond{97.8} & 90.0 & \liberoplussecond{93.5} & \liberoplusbest{88.5} & 82.1 \\

Cosmos-Policy~\cite{kim2026cosmospolicy}
& 75.8 & 63.3 & 81.7 & 96.5 & 88.9 & 92.7 & 82.2 & 82.2 \\

ImageWAM~\cite{zhang2026imagewam}
& \liberoplussecond{80.8} & 50.3 & \liberoplusbest{91.4} & \liberoplusbest{98.1} & 85.5 & \liberoplusbest{93.8} & 80.5 & 83.1 \\

ABot-M0.5~\cite{chen2026abotm05}
& 70.5 & \liberoplusbest{87.4} & 88.6 & 94.0 & 89.7 & 75.5 & \liberoplussecond{85.2} & \liberoplussecond{83.4} \\

\midrule

\ours
& 72.9 & \liberoplussecond{71.9} & \liberoplussecond{90.3} & 97.5 & 90.4 & 88.0 & 82.7 & \liberoplusbest{84.0} \\

\bottomrule
\end{tabular}%
}
}
\end{table}

\noindent\textbf{LIBERO.}
LIBERO~\cite{liu2023libero} is a simulated 7-DoF single-arm manipulation
benchmark comprising four suites: Spatial, Object, Goal, and Long.
Each suite contains 10 tasks and emphasizes a different aspect of
generalization.
We train a single multi-task policy on 1,693 trajectories spanning
40 tasks across the four suites.
Following the standard protocol, we evaluate each task over 50 episodes,
yielding 500 trials per suite and 2,000 trials overall.

\noindent\textbf{LIBERO-Plus.}
LIBERO-Plus~\cite{fei2026liberoplus} evaluates policy robustness under
distribution shifts in camera viewpoint, robot initial state, language,
illumination, background, sensor noise, and object layout.
We directly evaluate the policy trained on the original LIBERO data without
additional fine-tuning or adaptation.
Following the official protocol, each perturbation instance is evaluated once.
The complete evaluation contains 10,030 instances across the four LIBERO suites.
We report success rates for each perturbation type and the overall fraction of
successful rollouts.

\noindent\textbf{RoboTwin~2.0.}
RoboTwin~\cite{chen2025robotwin} is a large-scale bimanual manipulation benchmark covering diverse objects, interactions, and task horizons.
The clean training set contains 2,500 demonstrations, with 50 demonstrations per task.
We train a single multi-task policy on this set and conduct 100 trials per evaluated task in each of the clean (Easy) and domain-randomized (Hard) settings.

\noindent\textbf{Real-world setup.}
We conduct real-world experiments on the CoBoT Magic platform, which is built on the Mobile ALOHA system design~\cite{fu2024mobile}.
We evaluate three tasks: \textit{1) Table Cleaning}, \textit{2) Occluded Placement}, and \textit{3) Bimanual Plug Insertion}.
These tasks respectively emphasize long-horizon execution, global-to-local coordination under occlusion, and fine-grained bimanual manipulation.
We collect 100 teleoperated trajectories per task, yielding 300 trajectories in total.
Detailed task configurations are provided in Appendix~\ref{app:real-world-details}.

\subsection{Simulation Results}
\noindent\textbf{LIBERO.}
Table~\ref{table:libero_comparison} compares \ours with generalist,
future-predictive, and reasoning-enhanced VLA baselines.
\ours achieves the best average success rate of $98.5\%$, exceeding the
strongest baseline by $1.3$ percentage points.
It obtains the highest success rates on Spatial, Object, and Goal, reaching
$99.6\%$, $99.8\%$, and $99.2\%$, respectively.
On the more temporally extended Long suite, \ours reaches $95.2\%$ and remains
competitive with the strongest reported methods.
These results show that task-conditioned future-wrist modeling improves
performance without sacrificing broad task generalization.

\noindent\textbf{LIBERO-Plus.}
Table~\ref{table:libero_plus} compares zero-shot robustness under seven
distribution shifts.
\ours achieves the highest overall success rate of $84.0\%$ across all 10,030
perturbation instances.
This result exceeds ABot-M0.5, the strongest baseline in overall performance,
by $0.6$ percentage points.
\ours also ranks second under robot-state and language perturbations, reaching
$71.9\%$ and $90.3\%$, respectively.
The remaining category results are consistently strong, indicating balanced
robustness rather than dependence on a single perturbation type.
These results indicate that task-conditioned future-wrist modeling supports robust
zero-shot generalization beyond the training distribution of the original LIBERO benchmark.

\noindent\textbf{RoboTwin~2.0.}
Table~\ref{tab:main_results} reports the average success rates on
RoboTwin~2.0 under the Easy and Hard settings.
Under the clean (Easy) setting, \ours achieves $60.71\%$, outperforming the strongest baseline, UP-VLA, by $7.79$ percentage points.
It also exceeds StarVLA-OFT and StarVLA-GR00T by $10.33$ and $11.91$
percentage points, respectively.
Under the domain-randomized (Hard) setting, \ours achieves $18.21\%$,
surpassing the strongest reported baseline, $\pi_0$, by $1.87$ percentage points and UP-VLA by $3.05$ percentage points.
These gains under both clean and domain-randomized settings support the value of conditioning wrist-local prediction on the current task context.

\begin{table}[!t]
\newcommand{\robotwinbest}[1]{\liberocaptionbox{linecolor1}{\textbf{#1}}}
\newcommand{\robotwinsecond}[1]{\liberocaptionbox{linecolor2}{\underline{#1}}}

\centering

\caption{
\textbf{Average success rates (\%) on RoboTwin~2.0 under the clean
(Easy) and domain-randomized (Hard) settings.}
\protect\liberocaptionbox{linecolor1}{\textbf{Bold}} denotes the best performance, and
\protect\liberocaptionbox{linecolor2}{\underline{underline}} denotes the second best.
}
\label{tab:main_results}

{
\small
\setlength{\tabcolsep}{4pt}
\renewcommand{\arraystretch}{1.12}

\begin{tabularx}{\linewidth}{
    @{}
    >{\raggedright\arraybackslash}X
    >{\centering\arraybackslash}p{0.18\linewidth}
    >{\centering\arraybackslash}p{0.18\linewidth}
    @{}
}
    \toprule
    \textbf{Method} & \textbf{Easy} & \textbf{Hard} \\
    \midrule
    $\pi_0$~\cite{black2024pi_0}
        & 46.42
        & \robotwinsecond{16.34} \\
    RDT~\cite{liu2025rdt}
        & 34.50
        & 13.72 \\
    Diffusion Policy~\cite{chi2025diffusion}
        & 28.06
        & 0.64 \\
    UP-VLA~\cite{zhang2025up}
        & \robotwinsecond{52.92}
        & 15.16 \\
    StarVLA-OFT~\cite{community2026starvla}
        & 50.38
        & -- \\
    StarVLA-GR00T~\cite{community2026starvla}
        & 48.80
        & -- \\
    \ours
        & \robotwinbest{60.71}
        & \robotwinbest{18.21} \\
    \bottomrule
\end{tabularx}
}
\end{table}

\subsection{Real-World Evaluation}
\label{sec:real_world}

\noindent\textbf{Evaluation protocol.}
We compare \ours with $\pi_0$ and VLA-JEPA, each fine-tuned on the same demonstration data and evaluated under the same protocol.
For each task and method, we conduct 30 trials under standard conditions and 30 OOD trials, evenly split across \textit{1) table clutter}, \textit{2) random lighting perturbations}, and \textit{3) background variations}.
We report binary task success and a stage-wise progress score that counts completed stages in the predefined task sequence.
The maximum scores $R$ are 4, 3, and 3 for \textit{Table Cleaning}, \textit{Occluded Placement}, and \textit{Bimanual Plug Insertion}, respectively.
The task-specific progress scoring rubric is summarized in
Table~\ref{tab:real-world-progress-rubric}.

\begin{table}[t]
    \centering

    \caption{\textbf{Real-world Task Progress-score rubric.} The parentheses show (progress/total score)}
    \label{tab:real-world-progress-rubric}

    {
    \small
    \setlength{\tabcolsep}{3pt}

    \begin{tabularx}{\columnwidth}{
        @{}
        >{\raggedright\arraybackslash}p{0.28\columnwidth}
        >{\raggedright\arraybackslash}X
        @{}
    }
        \toprule
        \textbf{Task} & \textbf{Ordered subtasks} \\
        \midrule
        Table Cleaning (\(R=4\))
        & \textbf{(1/4)} left paper ball to basket;
          \textbf{(2/4)} left one block to basket;
          \textbf{(3/4)} right other block to basket;
          \textbf{(4/4)} right wipe spill with cloth. \\
        \addlinespace
        Occluded Placement (\(R=3\))
        & \textbf{(1/3)} approach mango without hitting obstacle;
          \textbf{(2/3)} grasp and move mango;
          \textbf{(3/3)} place mango on plate. \\
        \addlinespace
        Bimanual Plug Insertion (\(R=3\))
        & \textbf{(1/3)} left: hold power strip;
          \textbf{(2/3)} right: grasp plug;
          \textbf{(3/3)} right: insert plug. \\
        \bottomrule
    \end{tabularx}
    }
\end{table}

\begin{figure}[!htbp]
\centering
\includegraphics[width=\linewidth]{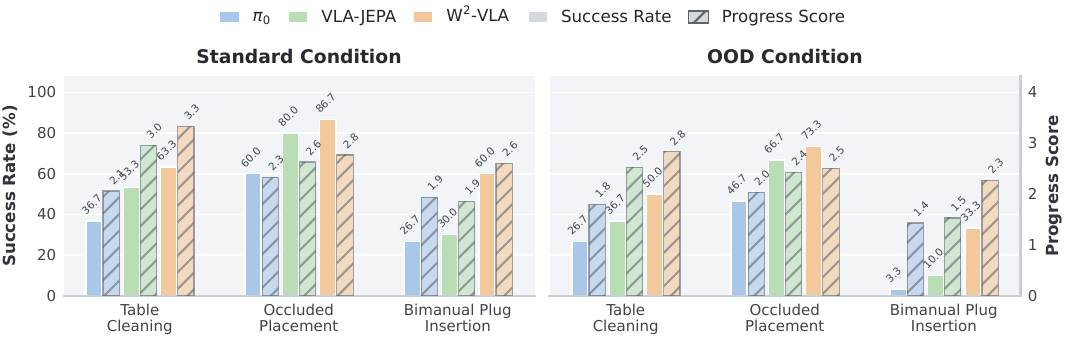}
\caption{Real-world performance under standard and OOD conditions.
Solid bars report success rates using the left axis, while hatched bars
report average progress scores using the right axis.
Higher values indicate better performance for both metrics shown in the figure.}
\label{fig:real_results}
\end{figure}

\begin{figure}[!htbp]
\centering
\includegraphics[width=\linewidth]{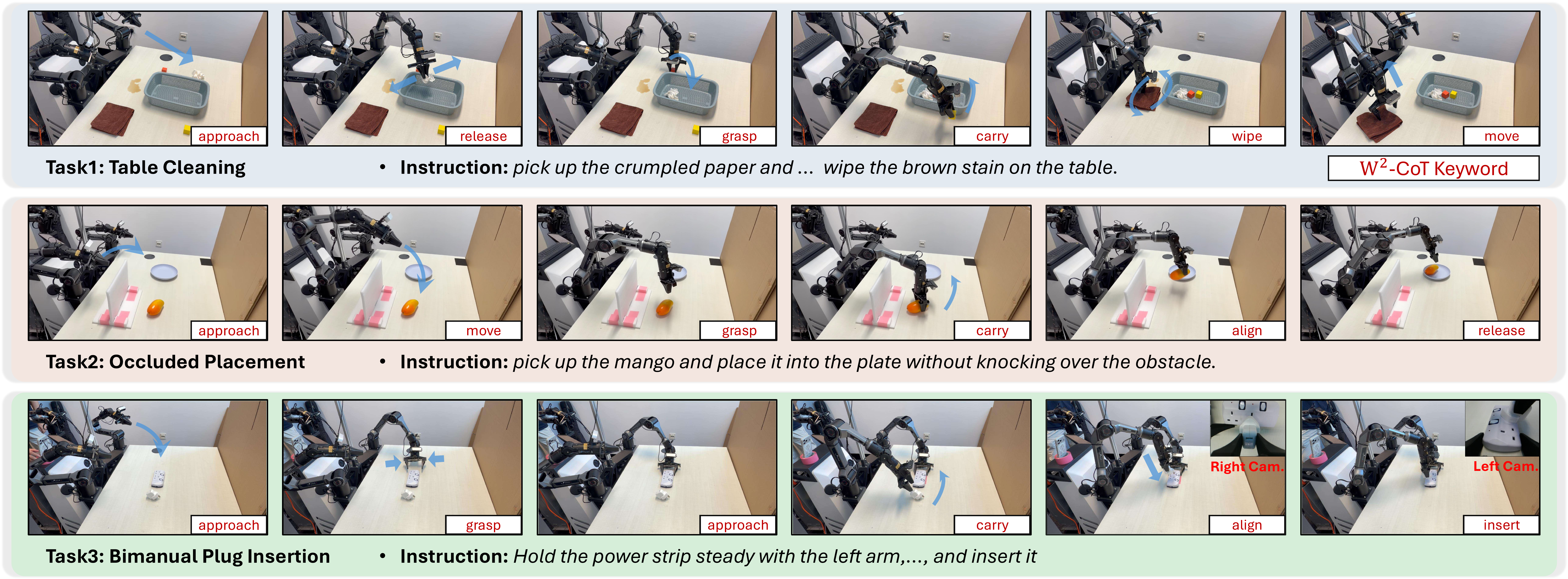}
\caption{\textbf{Real-world task rollouts.}
\textit{\#1 Table Cleaning} requires long-horizon object collection and
wiping.
\textit{\#2 Occluded Placement} requires the robot to route around an
obstacle before local grasping and placement.
\textit{\#3 Bimanual Plug Insertion} requires fine-grained action, one arm to stabilize the power
strip while the other aligns and inserts the plug.}
\label{fig:rollouts}
\end{figure}

\noindent\textbf{Standard and OOD performance.}
Fig.~\ref{fig:real_results} summarizes performance across the three tasks.
Under standard conditions, \ours achieves the highest success rate on all
three tasks, averaging $70.00\%$ and outperforming VLA-JEPA and $\pi_0$ by
$15.56$ and $28.89$ percentage points, respectively.
Under OOD conditions, \ours continues to achieve the highest success rate on
all three tasks, with an average of $52.22\%$, exceeding VLA-JEPA by $14.44$
percentage points.
The largest OOD margin over VLA-JEPA occurs on Bimanual Plug Insertion, where \ours achieves $33.33\%$ success, compared with $10.00\%$ for VLA-JEPA.
Representative rollouts in Fig.~\ref{fig:rollouts} illustrate long-horizon
execution, manipulation under occlusion, and fine-grained interactions
across these tasks.
During real-world deployment, \ours generates a 16-step action chunk in 183\,ms, yielding an action-generation rate of 87.43\,Hz and supporting real-time deployment.

\noindent\textbf{Progress and failure analysis.}
\ours also achieves the highest progress score on every task under both
evaluation settings, indicating more reliable partial completion even when
the full task is not completed.
On Bimanual Plug Insertion, \ours improves the progress score over VLA-JEPA
from $1.86$ to $2.60$ under standard conditions and from $1.53$ to $2.27$
under OOD conditions.
Failure inspection on this task shows that the evaluated policies often
complete the initial grasping stages but fail during final alignment or
insertion.
The higher progress scores show that \ours reaches these contact-sensitive
stages more frequently, even when full completion remains difficult.

\begin{table}[!t]
\centering

\caption{\textbf{Ablation studies on LIBERO.}
(a) Component contributions.
(b) Interface design and inference latency.
(c) Future-view prediction targets.
Average success is reported across four suites; bold denotes the best result in each panel.}
\label{table:ablation}

{
\footnotesize
\setlength{\tabcolsep}{2pt}
\renewcommand{\arraystretch}{1.10}

\begin{tabular*}{0.94\columnwidth}{@{\extracolsep{\fill}}lccccc@{}}
\toprule
\multicolumn{6}{c}{\textit{(a) Component contributions}} \\
\textbf{Configuration} &
\shortstack{\textbf{Spatial}\\\textbf{(\%)}} &
\shortstack{\textbf{Object}\\\textbf{(\%)}} &
\shortstack{\textbf{Goal}\\\textbf{(\%)}} &
\shortstack{\textbf{Long}\\\textbf{(\%)}} &
\shortstack{\textbf{Avg.}\\\textbf{(\%)}} \\
\midrule
\ours
    & \textbf{99.6}
    & \textbf{99.8}
    & \textbf{99.2}
    & \textbf{95.2}
    & \textbf{98.5} \\
w/o Wrist Predictor
    & 98.6
    & 99.6
    & 98.2
    & 93.6
    & 97.5 \\
w/o \CoT
    & 99.0
    & 99.2
    & 98.8
    & 95.0
    & 98.0 \\
    \bottomrule
\end{tabular*}

\vspace{1.3mm}

\begin{tabular*}{0.94\columnwidth}{@{\extracolsep{\fill}}cccccc@{}}
\multicolumn{6}{c}{
    \textit{(b) Interface design and inference efficiency}
} \\
\textbf{Idx.} &
\shortstack{\textbf{Decode CoT}\\\textbf{at Inference}} &
\shortstack{\textbf{Wrist}\\\textbf{Predictor}} &
\shortstack{\textbf{Latent}\\\textbf{Tokens}} &
\shortstack{\textbf{Latency}\\\textbf{(ms)}} &
\shortstack{\textbf{Avg.}\\\textbf{(\%)}} \\
\midrule
1
    & \Checkmark
    & \XSolidBrush
    & N/A
    & 1550.77
    & 97.6 \\
2
    & \Checkmark
    & \Checkmark
    & N/A
    & 1615.27
    & 98.1 \\
3
    & \XSolidBrush
    & \Checkmark
    & 4
    & \textbf{98.58}
    & 98.0 \\
4
    & \XSolidBrush
    & \Checkmark
    & 8
    & 102.15
    & 98.1 \\
5
    & \XSolidBrush
    & \Checkmark
    & 32
    & 148.69
    & 98.4 \\
\addlinespace[0.25em]
\ours
    & \XSolidBrush
    & \Checkmark
    & 16
    & 110.58
    & \textbf{98.5} \\
    \bottomrule
\end{tabular*}

\vspace{1.3mm}

\begin{tabular*}{0.94\columnwidth}{@{\extracolsep{\fill}}ccccc@{}}
\multicolumn{5}{c}{\textit{(c) Future-view prediction targets}} \\
\textbf{Idx.} &
\shortstack{\textbf{Main-view}\\\textbf{Prediction}} &
\shortstack{\textbf{Wrist-view}\\\textbf{Prediction}} &
\shortstack{\textbf{Latency}\\\textbf{(ms)}} &
\shortstack{\textbf{Avg.}\\\textbf{(\%)}} \\
\midrule
1
    & \Checkmark
    & \XSolidBrush
    & \textbf{102.49}
    & 97.7 \\
2
    & \Checkmark
    & \Checkmark
    & 132.76
    & 98.0 \\
\addlinespace[0.25em]
\ours
    & \XSolidBrush
    & \Checkmark
    & 110.58
    & \textbf{98.5} \\
\bottomrule
\end{tabular*}
}
\end{table}

\subsection{Ablation Studies}

\noindent\textbf{Component contributions.}
Table~\ref{table:ablation}(a) examines the individual contributions of
structured \CoT supervision and future-wrist prediction.
Removing the Wrist Predictor lowers the average success rate from $98.5\%$ to
$97.5\%$.
The largest decrease occurs on Long, where performance drops from $95.2\%$ to
$93.6\%$, suggesting that future wrist prediction is especially useful for
temporally extended manipulation.
Removing \CoT supervision reduces the average success rate to $98.0\%$.
Structured annotation prediction helps shape the task-conditioned interface,
whereas future-wrist prediction provides a localized predictive objective.
Their combination yields the strongest overall performance.

\noindent\textbf{Interface design and inference efficiency.}
We compare fixed latent modeling tokens with explicit CoT conditioning.
The explicit variants require autoregressive CoT decoding at inference; when
the Wrist Predictor is enabled, it is conditioned on the hidden states of the
decoded CoT.
As shown in Table~\ref{table:ablation}(b), explicit decoding requires more
than $1.5$ seconds per action chunk.
In contrast, the fixed latent interface reduces latency by more than an order of magnitude while achieving comparable or higher success rates.
Increasing the number of latent modeling tokens from 4 to 32 yields no
consistent improvement and gradually increases latency.
The 16-token configuration achieves the best average success rate of $98.5\%$ with a latency of $110.58$ ms.

\noindent\textbf{Future-view prediction targets.}
Finally, we examine whether future prediction should target the main view, the wrist view, or both views jointly.
Table~\ref{table:ablation}(c) shows that wrist-only future prediction achieves the strongest result.
The fixed main camera contains substantial static scene content, so its future
latents may place less emphasis on action-induced local changes.
Wrist observations instead capture evolving gripper--object geometry,
contact, alignment, and release.
Jointly predicting both views also increases computational cost and may introduce interference between their distinct prediction signals, diluting the action-relevant supervision from the wrist view.
Therefore, we restrict prediction to wrist views, focusing the predictive objective on local dynamics.

\subsection{Qualitative Analysis}
\label{sec:qualitative}
\begin{figure}[!htbp]
\centering
\includegraphics[width=\linewidth]{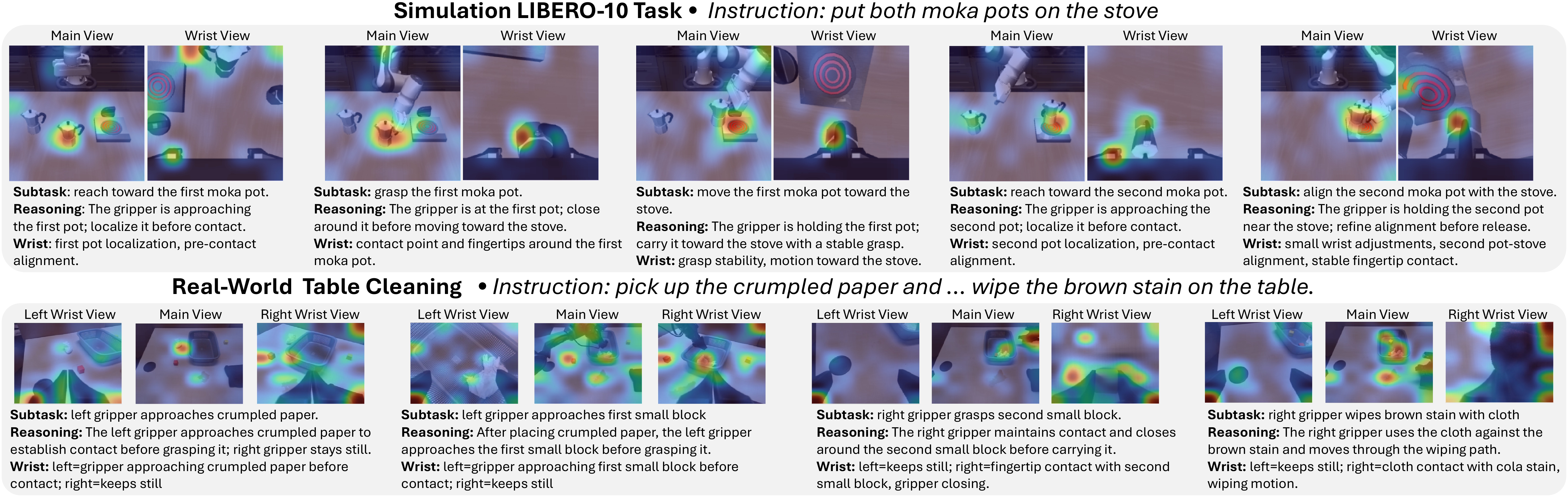}
\caption{\textbf{Latent modeling token attention over multi-view observations.}
Attention maps are shown over main- and wrist-view image tokens across
manipulation stages.
For semantic interpretation, the VLM language head decodes structured
\texttt{Subtask}/\texttt{Reasoning}/\texttt{Wrist} descriptions; the CoT decoding
is only for visualization.}
\label{fig:visual}
\end{figure}

\noindent\textbf{Attention visualization.}
Fig.~\ref{fig:visual} shows that the latent modeling tokens attend to stage-relevant regions in the main and wrist views.
We extract direct post-softmax self-attention weights from the final language-transformer layer and average them over all attention heads and latent modeling tokens.
Their attention shifts among target objects, grippers, and contact areas as manipulation progresses.
In LIBERO-10, attention moves from graspable regions during approach and grasp to the held object and target area during transport and alignment.
In the Table Cleaning task, it follows the active object and gripper before concentrating on cloth--stain contact during wiping.
These patterns suggest that the task-conditioned interface draws on manipulation-relevant global and wrist-local evidence when conditioning future wrist prediction.
See more visualizations in Appendix~\ref{appendix:c}.

\FloatBarrier
\section{Conclusion}

We presented \ours, a task-conditioned future wrist modeling framework for fine-grained robot manipulation.
Rather than treating main- and wrist-view observations as parallel inputs, \ours uses a compact latent interface shaped by structured \CoT supervision to connect global task context with wrist-local prediction.
Conditioned on this interface and wrist history, a predictor forecasts future wrist latents as future-aware context for action generation.
Experiments on LIBERO, RoboTwin~2.0, and real-world tasks demonstrate strong performance across single-arm and bimanual settings.
The policy does not require CoT generation at inference time, enabling real-time action generation at over 80 Hz.

\newpage
\bibliography{references}

@misc{brohan2022rt1,
      title={RT-1: Robotics Transformer for Real-World Control at Scale}, 
      author={Anthony Brohan and Noah Brown and Justice Carbajal and Yevgen Chebotar and Joseph Dabis and Chelsea Finn and Keerthana Gopalakrishnan and Karol Hausman and Alex Herzog and Jasmine Hsu and Julian Ibarz and Brian Ichter and Alex Irpan and Tomas Jackson and Sally Jesmonth and Nikhil J Joshi and Ryan Julian and Dmitry Kalashnikov and Yuheng Kuang and Isabel Leal and Kuang-Huei Lee and Sergey Levine and Yao Lu and Utsav Malla and Deeksha Manjunath and Igor Mordatch and Ofir Nachum and Carolina Parada and Jodilyn Peralta and Emily Perez and Karl Pertsch and Jornell Quiambao and Kanishka Rao and Michael Ryoo and Grecia Salazar and Pannag Sanketi and Kevin Sayed and Jaspiar Singh and Sumedh Sontakke and Austin Stone and Clayton Tan and Huong Tran and Vincent Vanhoucke and Steve Vega and Quan Vuong and Fei Xia and Ted Xiao and Peng Xu and Sichun Xu and Tianhe Yu and Brianna Zitkovich},
      year={2023},
      eprint={2212.06817},
      archivePrefix={arXiv},
      primaryClass={cs.RO},
      url={https://arxiv.org/abs/2212.06817}, 
}

@inproceedings{zitkovich2023rt,
  title={Rt-2: Vision-language-action models transfer web knowledge to robotic control},
  author={Zitkovich, Brianna and Yu, Tianhe and Xu, Sichun and Xu, Peng and Xiao, Ted and Xia, Fei and Wu, Jialin and Wohlhart, Paul and Welker, Stefan and Wahid, Ayzaan and others},
  booktitle={Conference on Robot Learning},
  pages={2165--2183},
  year={2023},
  organization={PMLR}
}

@inproceedings{kim2024openvla,
  title={OpenVLA: An Open-Source Vision-Language-Action Model},
  author={Kim, Moo Jin and Pertsch, Karl and Karamcheti, Siddharth and Xiao, Ted and Balakrishna, Ashwin and Nair, Suraj and Rafailov, Rafael and Foster, Ethan P and Sanketi, Pannag R and Vuong, Quan and others},
  booktitle={Conference on Robot Learning},
  pages={2679--2713},
  year={2025},
  organization={PMLR}
}

@inproceedings{octo2024,
    title={Octo: An Open-Source Generalist Robot Policy},
    author = {{Octo Model Team} and Dibya Ghosh and Homer Walke and Karl Pertsch and Kevin Black and Oier Mees and Sudeep Dasari and Joey Hejna and Charles Xu and Jianlan Luo and Tobias Kreiman and {You Liang} Tan and Lawrence Yunliang Chen and Pannag Sanketi and Quan Vuong and Ted Xiao and Dorsa Sadigh and Chelsea Finn and Sergey Levine},
    booktitle = {Proceedings of Robotics: Science and Systems},
    address  = {Delft, Netherlands},
    year = {2024},
}

@misc{black2024pi_0,
      title={$\pi_0$: A Vision-Language-Action Flow Model for General Robot Control}, 
      author={Kevin Black and Noah Brown and Danny Driess and Adnan Esmail and Michael Equi and Chelsea Finn and Niccolo Fusai and Lachy Groom and Karol Hausman and Brian Ichter and Szymon Jakubczak and Tim Jones and Liyiming Ke and Sergey Levine and Adrian Li-Bell and Mohith Mothukuri and Suraj Nair and Karl Pertsch and Lucy Xiaoyang Shi and James Tanner and Quan Vuong and Anna Walling and Haohuan Wang and Ury Zhilinsky},
      year={2026},
      eprint={2410.24164},
      archivePrefix={arXiv},
      primaryClass={cs.LG},
      url={https://arxiv.org/abs/2410.24164}, 
}

@misc{qwen3vl2025,
      title={Qwen3-VL Technical Report}, 
      author={Shuai Bai and Yuxuan Cai and Ruizhe Chen and Keqin Chen and Xionghui Chen and Zesen Cheng and Lianghao Deng and Wei Ding and Chang Gao and Chunjiang Ge and Wenbin Ge and Zhifang Guo and Qidong Huang and Jie Huang and Fei Huang and Binyuan Hui and Shutong Jiang and Zhaohai Li and Mingsheng Li and Mei Li and Kaixin Li and Zicheng Lin and Junyang Lin and Xuejing Liu and Jiawei Liu and Chenglong Liu and Yang Liu and Dayiheng Liu and Shixuan Liu and Dunjie Lu and Ruilin Luo and Chenxu Lv and Rui Men and Lingchen Meng and Xuancheng Ren and Xingzhang Ren and Sibo Song and Yuchong Sun and Jun Tang and Jianhong Tu and Jianqiang Wan and Peng Wang and Pengfei Wang and Qiuyue Wang and Yuxuan Wang and Tianbao Xie and Yiheng Xu and Haiyang Xu and Jin Xu and Zhibo Yang and Mingkun Yang and Jianxin Yang and An Yang and Bowen Yu and Fei Zhang and Hang Zhang and Xi Zhang and Bo Zheng and Humen Zhong and Jingren Zhou and Fan Zhou and Jing Zhou and Yuanzhi Zhu and Ke Zhu},
      year={2025},
      eprint={2511.21631},
      archivePrefix={arXiv},
      primaryClass={cs.CV},
      url={https://arxiv.org/abs/2511.21631}, 
}

@misc{mur2026v,
      title={V-JEPA 2.1: Unlocking Dense Features in Video Self-Supervised Learning}, 
      author={Lorenzo Mur-Labadia and Matthew Muckley and Amir Bar and Mido Assran and Koustuv Sinha and Mike Rabbat and Yann LeCun and Nicolas Ballas and Adrien Bardes},
      year={2026},
      eprint={2603.14482},
      archivePrefix={arXiv},
      primaryClass={cs.CV},
      url={https://arxiv.org/abs/2603.14482}, 
}

@inproceedings{peebles2023scalable,
  title={Scalable diffusion models with transformers},
  author={Peebles, William and Xie, Saining},
  booktitle={Proceedings of the IEEE/CVF international conference on computer vision},
  pages={4195--4205},
  year={2023}
}

@misc{qu2025spatialvla,
      title={SpatialVLA: Exploring Spatial Representations for Visual-Language-Action Model}, 
      author={Delin Qu and Haoming Song and Qizhi Chen and Yuanqi Yao and Xinyi Ye and Yan Ding and Zhigang Wang and JiaYuan Gu and Bin Zhao and Dong Wang and Xuelong Li},
      year={2025},
      eprint={2501.15830},
      archivePrefix={arXiv},
      primaryClass={cs.RO},
      url={https://arxiv.org/abs/2501.15830}, 
}

@misc{spatialforcing2025,
      title={Spatial Forcing: Implicit Spatial Representation Alignment for Vision-language-action Model}, 
      author={Fuhao Li and Wenxuan Song and Han Zhao and Jingbo Wang and Pengxiang Ding and Donglin Wang and Long Zeng and Haoang Li},
      year={2025},
      eprint={2510.12276},
      archivePrefix={arXiv},
      primaryClass={cs.RO},
      url={https://arxiv.org/abs/2510.12276}, 
}

@inproceedings{huang2026graphcot,
  title={Graphcot-vla: A 3d spatial-aware reasoning vision-language-action model for robotic manipulation with ambiguous instructions},
  author={Huang, Helong and Cen, Min and Tan, Kai and Quan, Xingyue and Huang, Guowei and Zhang, Hong},
  booktitle={Proceedings of the AAAI Conference on Artificial Intelligence},
  volume={40},
  pages={18324--18332},
  year={2026}
}

@misc{luo2026being,
      title={Being-H0.7: A Latent World-Action Model from Egocentric Videos}, 
      author={Hao Luo and Wanpeng Zhang and Yicheng Feng and Sipeng Zheng and Haiweng Xu and Chaoyi Xu and Ziheng Xi and Yuhui Fu and Zongqing Lu},
      year={2026},
      eprint={2605.00078},
      archivePrefix={arXiv},
      primaryClass={cs.RO},
      url={https://arxiv.org/abs/2605.00078}, 
}

@inproceedings{
shouhalo,
title={{HALO}: A Unified Vision-Language-Action Model for Embodied Multimodal Chain-of-Thought Reasoning},
author={Quanxin Shou and Fangqi Zhu and Shuang Chen and Puxin Yan and Zhengyang Yan and Yikun Miao and Xiaoyi Pang and Zicong Hong and Ruikai Shi and Hao HUANG and Jie ZHANG and Song Guo},
booktitle={Forty-third International Conference on Machine Learning},
year={2026},
url={https://openreview.net/forum?id=lduY9csXqw}
}

@misc{bai2026latent,
      title={Latent Reasoning VLA: Latent Thinking and Prediction for Vision-Language-Action Models}, 
      author={Shuanghao Bai and Jing Lyu and Wanqi Zhou and Zhe Li and Dakai Wang and Lei Xing and Xiaoguang Zhao and Pengwei Wang and Zhongyuan Wang and Cheng Chi and Badong Chen and Shanghang Zhang},
      year={2026},
      eprint={2602.01166},
      archivePrefix={arXiv},
      primaryClass={cs.RO},
      url={https://arxiv.org/abs/2602.01166}, 
}

@misc{sun2026vla,
      title={VLA-JEPA: Enhancing Vision-Language-Action Model with Latent World Model}, 
      author={Jingwen Sun and Wenyao Zhang and Zekun Qi and Shaojie Ren and Zezhi Liu and Hanxin Zhu and Guangzhong Sun and Xin Jin and Zhibo Chen},
      year={2026},
      eprint={2602.10098},
      archivePrefix={arXiv},
      primaryClass={cs.RO},
      url={https://arxiv.org/abs/2602.10098}, 
}

@misc{zhang2026dreamvla,
      title={DreamVLA: A Vision-Language-Action Model Dreamed with Comprehensive World Knowledge}, 
      author={Wenyao Zhang and Hongsi Liu and Zekun Qi and Yunnan Wang and Xinqiang Yu and Jiazhao Zhang and Runpei Dong and Jiawei He and Fan Lu and He Wang and Zhizheng Zhang and Li Yi and Wenjun Zeng and Xin Jin},
      year={2025},
      eprint={2507.04447},
      archivePrefix={arXiv},
      primaryClass={cs.CV},
      url={https://arxiv.org/abs/2507.04447}, 
}

@misc{community2026starvla,
      title={StarVLA: A Lego-like Codebase for Vision-Language-Action Model Developing}, 
      author={StarVLA Community},
      year={2026},
      eprint={2604.05014},
      archivePrefix={arXiv},
      primaryClass={cs.RO},
      url={https://arxiv.org/abs/2604.05014}, 
}

@misc{intelligence2025pi_,
      title={$\pi_{0.5}$: a Vision-Language-Action Model with Open-World Generalization}, 
      author={Physical Intelligence and Kevin Black and Noah Brown and James Darpinian and Karan Dhabalia and Danny Driess and Adnan Esmail and Michael Equi and Chelsea Finn and Niccolo Fusai and Manuel Y. Galliker and Dibya Ghosh and Lachy Groom and Karol Hausman and Brian Ichter and Szymon Jakubczak and Tim Jones and Liyiming Ke and Devin LeBlanc and Sergey Levine and Adrian Li-Bell and Mohith Mothukuri and Suraj Nair and Karl Pertsch and Allen Z. Ren and Lucy Xiaoyang Shi and Laura Smith and Jost Tobias Springenberg and Kyle Stachowicz and James Tanner and Quan Vuong and Homer Walke and Anna Walling and Haohuan Wang and Lili Yu and Ury Zhilinsky},
      year={2025},
      eprint={2504.16054},
      archivePrefix={arXiv},
      primaryClass={cs.LG},
      url={https://arxiv.org/abs/2504.16054}, 
}

@misc{bjorck2025gr00t,
      title={GR00T N1: An Open Foundation Model for Generalist Humanoid Robots}, 
      author={NVIDIA and : and Johan Bjorck and Fernando Castañeda and Nikita Cherniadev and Xingye Da and Runyu Ding and Linxi "Jim" Fan and Yu Fang and Dieter Fox and Fengyuan Hu and Spencer Huang and Joel Jang and Zhenyu Jiang and Jan Kautz and Kaushil Kundalia and Lawrence Lao and Zhiqi Li and Zongyu Lin and Kevin Lin and Guilin Liu and Edith Llontop and Loic Magne and Ajay Mandlekar and Avnish Narayan and Soroush Nasiriany and Scott Reed and You Liang Tan and Guanzhi Wang and Zu Wang and Jing Wang and Qi Wang and Jiannan Xiang and Yuqi Xie and Yinzhen Xu and Zhenjia Xu and Seonghyeon Ye and Zhiding Yu and Ao Zhang and Hao Zhang and Yizhou Zhao and Ruijie Zheng and Yuke Zhu},
      year={2025},
      eprint={2503.14734},
      archivePrefix={arXiv},
      primaryClass={cs.RO},
      url={https://arxiv.org/abs/2503.14734}, 
}

@inproceedings{wen2025dexvla,
  title={DexVLA: Vision-Language Model with Plug-In Diffusion Expert for General Robot Control},
  author={Wen, Junjie and Zhu, Yichen and Li, Jinming and Tang, Zhibin and Shen, Chaomin and Feng, Feifei},
  booktitle={Conference on Robot Learning},
  pages={3094--3114},
  year={2025},
  organization={PMLR}
}

@misc{zheng2025x,
      title={X-VLA: Soft-Prompted Transformer as Scalable Cross-Embodiment Vision-Language-Action Model}, 
      author={Jinliang Zheng and Jianxiong Li and Zhihao Wang and Dongxiu Liu and Xirui Kang and Yuchun Feng and Yinan Zheng and Jiayin Zou and Yilun Chen and Jia Zeng and Ya-Qin Zhang and Jiangmiao Pang and Jingjing Liu and Tai Wang and Xianyuan Zhan},
      year={2025},
      eprint={2510.10274},
      archivePrefix={arXiv},
      primaryClass={cs.RO},
      url={https://arxiv.org/abs/2510.10274}, 
}

@inproceedings{wang2026vla,
  title={Vla-adapter: An effective paradigm for tiny-scale vision-language-action model},
  author={Wang, Yihao and Ding, Pengxiang and Li, Lingxiao and Cui, Can and Ge, Zirui and Tong, Xinyang and Song, Wenxuan and Zhao, Han and Zhao, Wei and Hou, Pengxu and others},
  booktitle={Proceedings of the AAAI conference on artificial intelligence},
  volume={40},
  pages={18638--18646},
  year={2026}
}

@inproceedings{zhao2025cot,
  title={Cot-vla: Visual chain-of-thought reasoning for vision-language-action models},
  author={Zhao, Qingqing and Lu, Yao and Kim, Moo Jin and Fu, Zipeng and Zhang, Zhuoyang and Wu, Yecheng and Li, Zhaoshuo and Ma, Qianli and Han, Song and Finn, Chelsea and others},
  booktitle={Proceedings of the Computer Vision and Pattern Recognition Conference},
  pages={1702--1713},
  year={2025}
}

@InProceedings{huang2026fast,
    author    = {Huang, Chi-Pin and Man, Yunze and Yu, Zhiding and Chen, Min-Hung and Kautz, Jan and Wang, Yu-Chiang Frank and Yang, Fu-En},
    title     = {Fast-ThinkAct: Efficient Vision-Language-Action Reasoning via Verbalizable Latent Planning},
    booktitle = {Proceedings of the IEEE/CVF Conference on Computer Vision and Pattern Recognition (CVPR)},
    month     = {June},
    year      = {2026},
    pages     = {5070-5081}
}

@misc{shi2025memoryvla,
      title={MemoryVLA: Perceptual-Cognitive Memory in Vision-Language-Action Models for Robotic Manipulation}, 
      author={Hao Shi and Bin Xie and Yingfei Liu and Lin Sun and Fengrong Liu and Tiancai Wang and Erjin Zhou and Haoqiang Fan and Xiangyu Zhang and Gao Huang},
      year={2026},
      eprint={2508.19236},
      archivePrefix={arXiv},
      primaryClass={cs.RO},
      url={https://arxiv.org/abs/2508.19236}, 
}

@inproceedings{zheng2025tracevla,
 author = {Zheng, Ruijie and Liang, Yongyuan and Huang, Shuaiyi and Gao, Jianfeng and Daum\'{e} III, Hal  and Kolobov, Andrey and Huang, Furong and Yang, Jianwei},
 booktitle = {International Conference on Learning Representations},
 editor = {Y. Yue and A. Garg and N. Peng and F. Sha and R. Yu},
 pages = {54277--54296},
 title = {TraceVLA: Visual Trace Prompting Enhances Spatial-Temporal Awareness for Generalist Robotic Policies},
 url = {https://proceedings.iclr.cc/paper_files/paper/2025/file/8667f264f88c7938a73a53ab01eb1327-Paper-Conference.pdf},
 volume = {2025},
 year = {2025}
}

@misc{yang2026hivla,
      title={HiVLA: A Visual-Grounded-Centric Hierarchical Embodied Manipulation System}, 
      author={Tianshuo Yang and Guanyu Chen and Yutian Chen and Zhixuan Liang and Yitian Liu and Zanxin Chen and Chunpu Xu and Haotian Liang and Jiangmiao Pang and Yao Mu and Ping Luo},
      year={2026},
      eprint={2604.14125},
      archivePrefix={arXiv},
      primaryClass={cs.CV},
      url={https://arxiv.org/abs/2604.14125}, 
}

@inproceedings{li2023blip,
  title={Blip-2: Bootstrapping language-image pre-training with frozen image encoders and large language models},
  author={Li, Junnan and Li, Dongxu and Savarese, Silvio and Hoi, Steven},
  booktitle={International conference on machine learning},
  pages={19730--19742},
  year={2023},
  organization={PMLR}
}

@article{liu2023libero,
  title={Libero: Benchmarking knowledge transfer for lifelong robot learning},
  author={Liu, Bo and Zhu, Yifeng and Gao, Chongkai and Feng, Yihao and Liu, Qiang and Zhu, Yuke and Stone, Peter},
  journal={Advances in Neural Information Processing Systems},
  volume={36},
  pages={44776--44791},
  year={2023}
}

@misc{chen2025robotwin,
      title={RoboTwin 2.0: A Scalable Data Generator and Benchmark with Strong Domain Randomization for Robust Bimanual Robotic Manipulation}, 
      author={Tianxing Chen and Zanxin Chen and Baijun Chen and Zijian Cai and Yibin Liu and Zixuan Li and Qiwei Liang and Xianliang Lin and Yiheng Ge and Zhenyu Gu and Weiliang Deng and Yubin Guo and Tian Nian and Xuanbing Xie and Qiangyu Chen and Kailun Su and Tianling Xu and Guodong Liu and Mengkang Hu and Huan-ang Gao and Kaixuan Wang and Zhixuan Liang and Yusen Qin and Xiaokang Yang and Ping Luo and Yao Mu},
      year={2025},
      eprint={2506.18088},
      archivePrefix={arXiv},
      primaryClass={cs.RO},
      url={https://arxiv.org/abs/2506.18088}, 
}

@misc{kim2025fine,
      title={Fine-Tuning Vision-Language-Action Models: Optimizing Speed and Success}, 
      author={Moo Jin Kim and Chelsea Finn and Percy Liang},
      year={2025},
      eprint={2502.19645},
      archivePrefix={arXiv},
      primaryClass={cs.RO},
      url={https://arxiv.org/abs/2502.19645}, 
}

@misc{pertsch2025fast,
      title={FAST: Efficient Action Tokenization for Vision-Language-Action Models}, 
      author={Karl Pertsch and Kyle Stachowicz and Brian Ichter and Danny Driess and Suraj Nair and Quan Vuong and Oier Mees and Chelsea Finn and Sergey Levine},
      year={2025},
      eprint={2501.09747},
      archivePrefix={arXiv},
      primaryClass={cs.RO},
      url={https://arxiv.org/abs/2501.09747}, 
}

@misc{yin2025deepthinkvla,
      title={DeepThinkVLA: Enhancing Reasoning Capability of Vision-Language-Action Models}, 
      author={Cheng Yin and Yankai Lin and Wang Xu and Sikyuen Tam and Xiangrui Zeng and Zhiyuan Liu and Zhouping Yin},
      year={2026},
      eprint={2511.15669},
      archivePrefix={arXiv},
      primaryClass={cs.LG},
      url={https://arxiv.org/abs/2511.15669}, 
}

@misc{fu2024mobile,
      title={Mobile ALOHA: Learning Bimanual Mobile Manipulation with Low-Cost Whole-Body Teleoperation}, 
      author={Zipeng Fu and Tony Z. Zhao and Chelsea Finn},
      year={2024},
      eprint={2401.02117},
      archivePrefix={arXiv},
      primaryClass={cs.RO},
      url={https://arxiv.org/abs/2401.02117}, 
}

@article{jangir2022look,
  title={Look closer: Bridging egocentric and third-person views with transformers for robotic manipulation},
  author={Jangir, Rishabh and Hansen, Nicklas and Ghosal, Sambaran and Jain, Mohit and Wang, Xiaolong},
  journal={IEEE Robotics and Automation Letters},
  volume={7},
  number={2},
  pages={3046--3053},
  year={2022},
  publisher={IEEE}
}

@article{lan2025bfa,
  title={Bfa: Best-feature-aware fusion for multi-view fine-grained manipulation},
  author={Lan, Zihan and Mao, Weixin and Li, Haosheng and Wang, Le and Wang, Tiancai and Fan, Haoqiang and Yoshie, Osamu},
  journal={IEEE Robotics and Automation Letters},
  year={2025},
  publisher={IEEE}
}

@misc{qian2025wristworld,
      title={WristWorld: Generating Wrist-Views via 4D World Models for Robotic Manipulation}, 
      author={Zezhong Qian and Xiaowei Chi and Yuming Li and Shizun Wang and Zhiyuan Qin and Xiaozhu Ju and Sirui Han and Shanghang Zhang},
      year={2025},
      eprint={2510.07313},
      archivePrefix={arXiv},
      primaryClass={cs.CV},
      url={https://arxiv.org/abs/2510.07313}, 
}

@misc{su2026world,
      title={World Guidance: World Modeling in Condition Space for Action Generation}, 
      author={Yue Su and Sijin Chen and Haixin Shi and Mingyu Liu and Zhengshen Zhang and Ningyuan Huang and Weiheng Zhong and Zhengbang Zhu and Yuxiao Liu and Xihui Liu},
      year={2026},
      eprint={2602.22010},
      archivePrefix={arXiv},
      primaryClass={cs.RO},
      url={https://arxiv.org/abs/2602.22010}, 
}

@misc{zhang2025spatial,
      title={From Spatial to Actions: Grounding Vision-Language-Action Model in Spatial Foundation Priors}, 
      author={Zhengshen Zhang and Hao Li and Yalun Dai and Zhengbang Zhu and Lei Zhou and Chenchen Liu and Dong Wang and Francis E. H. Tay and Sijin Chen and Ziwei Liu and Yuxiao Liu and Xinghang Li and Pan Zhou},
      year={2026},
      eprint={2510.17439},
      archivePrefix={arXiv},
      primaryClass={cs.RO},
      url={https://arxiv.org/abs/2510.17439}, 
}

@misc{li2026robointer,
      title={RoboInter: A Holistic Intermediate Representation Suite Towards Robotic Manipulation}, 
      author={Hao Li and Ziqin Wang and Zi-han Ding and Shuai Yang and Yilun Chen and Yang Tian and Xiaolin Hu and Tai Wang and Dahua Lin and Feng Zhao and Si Liu and Jiangmiao Pang},
      year={2026},
      eprint={2602.09973},
      archivePrefix={arXiv},
      primaryClass={cs.RO},
      url={https://arxiv.org/abs/2602.09973}, 
}

@misc{shukor2025smolvla,
      title={SmolVLA: A Vision-Language-Action Model for Affordable and Efficient Robotics}, 
      author={Mustafa Shukor and Dana Aubakirova and Francesco Capuano and Pepijn Kooijmans and Steven Palma and Adil Zouitine and Michel Aractingi and Caroline Pascal and Martino Russi and Andres Marafioti and Simon Alibert and Matthieu Cord and Thomas Wolf and Remi Cadene},
      year={2025},
      eprint={2506.01844},
      archivePrefix={arXiv},
      primaryClass={cs.LG},
      url={https://arxiv.org/abs/2506.01844}, 
}

@article{chi2025diffusion,
  title={Diffusion policy: Visuomotor policy learning via action diffusion},
  author={Chi, Cheng and Xu, Zhenjia and Feng, Siyuan and Cousineau, Eric and Du, Yilun and Burchfiel, Benjamin and Tedrake, Russ and Song, Shuran},
  journal={The International Journal of Robotics Research},
  volume={44},
  number={10-11},
  pages={1684--1704},
  year={2025},
  publisher={Sage Publications Sage UK: London, England}
}

@misc{zhang2025up,
      title={UP-VLA: A Unified Understanding and Prediction Model for Embodied Agent}, 
      author={Jianke Zhang and Yanjiang Guo and Yucheng Hu and Xiaoyu Chen and Xiang Zhu and Jianyu Chen},
      year={2025},
      eprint={2501.18867},
      archivePrefix={arXiv},
      primaryClass={cs.CV},
      url={https://arxiv.org/abs/2501.18867}, 
}

@misc{liu2025rdt,
      title={RDT-1B: a Diffusion Foundation Model for Bimanual Manipulation}, 
      author={Songming Liu and Lingxuan Wu and Bangguo Li and Hengkai Tan and Huayu Chen and Zhengyi Wang and Ke Xu and Hang Su and Jun Zhu},
      year={2025},
      eprint={2410.07864},
      archivePrefix={arXiv},
      primaryClass={cs.RO},
      url={https://arxiv.org/abs/2410.07864}, 
}

@misc{peng2025omnivggt,
      title={OmniVGGT: Omni-Modality Driven Visual Geometry Grounded Transformer}, 
      author={Haosong Peng and Hao Li and Yalun Dai and Yushi Lan and Yihang Luo and Tianyu Qi and Zhengshen Zhang and Yufeng Zhan and Junfei Zhang and Wenchao Xu and Ziwei Liu},
      year={2025},
      eprint={2511.10560},
      archivePrefix={arXiv},
      primaryClass={cs.CV},
      url={https://arxiv.org/abs/2511.10560}, 
}

@article{fei2026liberoplus,
  title={Libero-plus: In-depth robustness analysis of vision-language-action models},
  author={Fei, Senyu and Wang, Siyin and Shi, Junhao and Dai, Zihao and Cai, Jikun and Qian, Pengfang and Ji, Li and He, Xinzhe and Zhang, Shiduo and Fei, Zhaoye and others},
  journal={arXiv preprint arXiv:2510.13626},
  year={2025}
}

@inproceedings{yang2026mantis,
  title={Mantis: A versatile vision-language-action model with disentangled visual foresight},
  author={Yang, Yi and Li, Xueqi and Chen, Yiyang and Song, Jin and Wang, Yihan and Xiao, Zipeng and Su, Jiadi and Qiaoben, You and Liu, Pengfei and Deng, Zhijie},
  booktitle={Proceedings of the IEEE/CVF Conference on Computer Vision and Pattern Recognition},
  pages={42505--42515},
  year={2026}
}

@article{yuan2026fastwam,
  title={Fast-wam: Do world action models need test-time future imagination?},
  author={Yuan, Tianyuan and Dong, Zibin and Liu, Yicheng and Zhao, Hang},
  journal={arXiv preprint arXiv:2603.16666},
  year={2026}
}

@article{kim2026cosmospolicy,
  title={Cosmos policy: Fine-tuning video models for visuomotor control and planning},
  author={Kim, Moo Jin and Gao, Yihuai and Lin, Tsung-Yi and Lin, Yen-Chen and Ge, Yunhao and Lam, Grace and Liang, Percy and Song, Shuran and Liu, Ming-Yu and Finn, Chelsea and others},
  journal={arXiv preprint arXiv:2601.16163},
  year={2026}
}

@article{zhang2026imagewam,
  title={ImageWAM: Do World Action Models Really Need Video Generation, or Just Image Editing?},
  author={Zhang, Yuyang and Zhang, Wenyao and Qi, Zekun and Zhang, He and Lin, Haitao and Zhang, Jingbo and Mu, Yao and Yang, Xiaokang and Zeng, Wenjun and Jin, Xin},
  journal={arXiv preprint arXiv:2606.19531},
  year={2026}
}

@article{chen2026abotm05,
  title={Abot-m0. 5: Unified mobility-and-manipulation world action model},
  author={Chen, Ronghan and Yang, Yandan and Tang, Zuojin and Huo, Dongjie and Lin, Tong and Wu, Haoning and Liu, Haoyun and Chen, Yuzhi and Zheng, Lulu and Yuan, Botai and others},
  journal={arXiv preprint arXiv:2607.00678},
  year={2026}
}

\clearpage
\appendix
\addtocontents{toc}{\protect\setcounter{tocdepth}{2}}
\setcounter{tocdepth}{2}

\begingroup
\renewcommand{\contentsname}{Supplementary Material}
\makeatletter
\renewcommand*\l@subsection{\@dottedtocline{2}{1.5em}{2.5em}}
\makeatother
\tableofcontents
\endgroup
\clearpage

\begin{figure}[!t]
\centering
\includegraphics[width=1\textwidth]{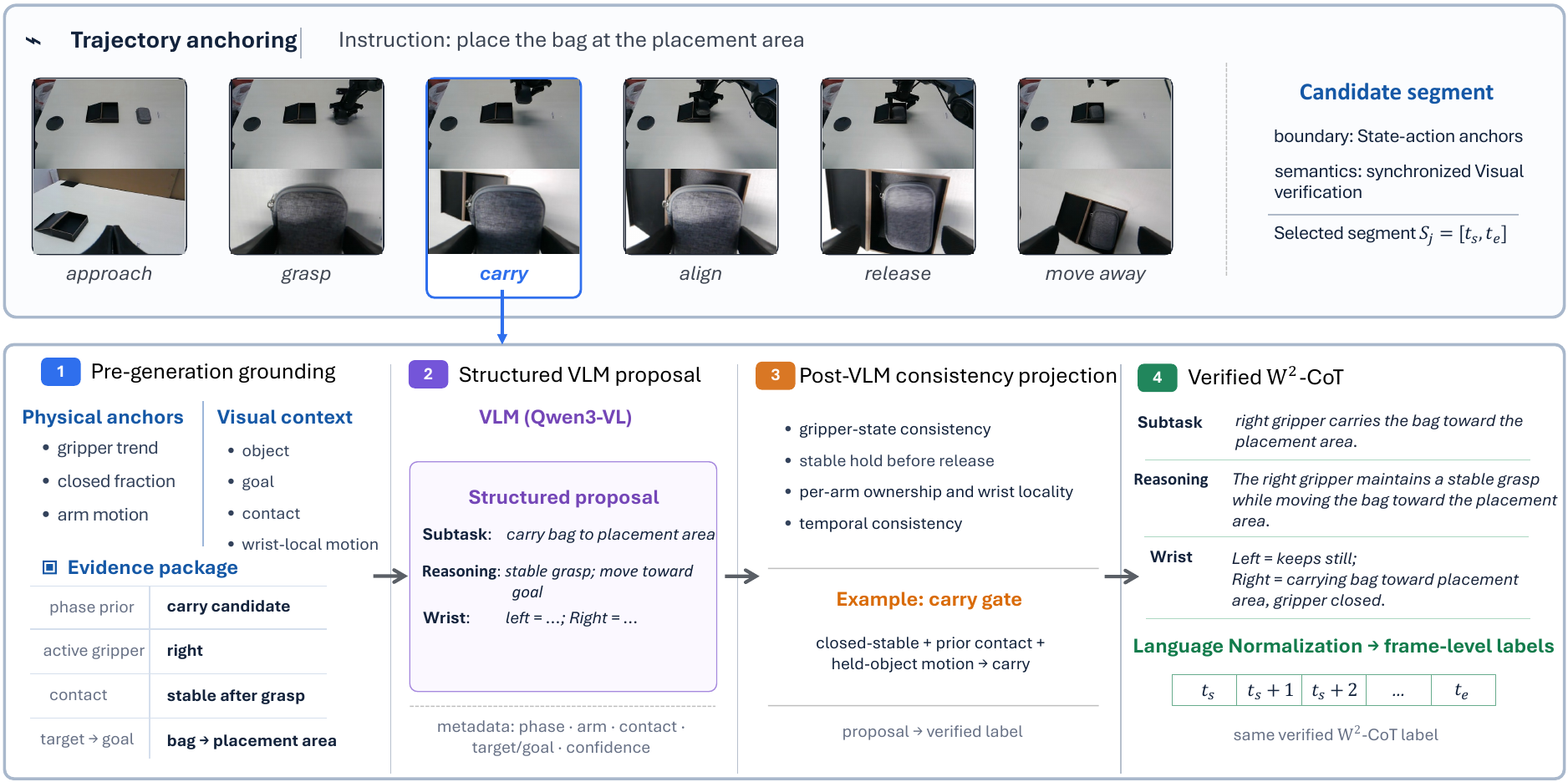}
\caption{\textbf{Overview of the \CoT annotation pipeline.}
State-action transitions and synchronized visual context first define candidate trajectory segments.
Each segment is grounded with physical and visual evidence before structured VLM proposal generation.
Deterministic consistency checks enforce gripper-state consistency, release preconditions, wrist locality, and temporal consistency.
The verified segment annotations are then language-normalized and expanded into frame-level \texttt{Subtask}, \texttt{Reasoning}, and \texttt{Wrist} supervision.}
\label{fig:cot_pipeline}
\end{figure}

\section{Details of the \texorpdfstring{\CoT}{W2-CoT} Annotation Pipeline}
\label{app:cot_construction}
Fig.~\ref{fig:cot_pipeline} summarizes the four-stage annotation pipeline.
The annotation pipeline converts raw robot trajectories into compact CoT labels through four stages.
The available state fields and camera views vary across datasets, but the construction principle remains unchanged.
We first parse robot states and actions to extract gripper openness, end-effector motion, and action changes. Gripper states indicate closure and release, while motion cues distinguish approach, transport, and post-release withdrawal.
Visual keyframes are sampled around candidate phase boundaries to verify object identity, contact, placement, and local motion.
The instruction, state evidence, and visual observations jointly form an evidence package for VLM-based CoT generation.

The VLM generates a structured CoT proposal conditioned on the evidence package. After physical verification and language normalization, the final supervision is rendered in three fields: \texttt{Subtask}, \texttt{Reasoning}, and \texttt{Wrist}.
The \texttt{Subtask} field describes the current \textit{manipulation progress}.
The \texttt{Reasoning} field explains \textit{physical transition cues}, that is, how this stage contributes to task progress under the observed state.
The \texttt{Wrist} field specifies \textit{wrist-local evidence}, including contact, object motion, placement stability, and gripper-object separation.
For bimanual trajectories, \texttt{Wrist} follows the ordered format \texttt{left=...; right=...}.

The generated proposal is then verified and corrected using physical consistency constraints.
An open gripper cannot be described as grasping or carrying an object.
A release stage must be supported by a preceding holding stage.
A carrying stage requires evidence of a stable grasp from the gripper state, object contact, or visual observations.
For bimanual trajectories, each wrist description must refer only to the corresponding gripper, object, and local motion. The temporal sequence is also checked to prevent physically implausible phase regressions. Short jitter and retry motions are prevented from introducing spurious task-level stages. Post-release withdrawal is retained only when it represents a sustained and physically meaningful motion.

Finally, the annotations are normalized into a consistent language style. Approach descriptions specify an open gripper moving toward the target when supported by the physical state.
Grasp descriptions emphasize fingertip contact and gripper closure.
Transport descriptions specify the carried object and its destination.
Release descriptions capture gripper opening, object stability, and gripper-object separation.
Post-release withdrawal is described as an open gripper moving away from the placed object. This normalization reduces linguistic variation while preserving the physical meaning of each stage.

\subsection{Representative Generation and Verification Templates}
Table~\ref{tab:cot-prompt-templates} summarizes two representative annotation interfaces used across the three evaluation settings. The wording is condensed for presentation rather than copied verbatim from the source code.
LIBERO uses episode-level planning with segment-wise visual grounding, whereas RoboTwin uses contact-sheet-based, state-anchored bimanual adjudication.
The real-world annotations follow the RoboTwin-style interface with task-specific physical constraints.
The deterministic post-generation projection is listed separately from the VLM prompts.

\begin{table}[!t]
\centering

\caption{\textbf{Condensed generation and verification templates for \CoT synthesis. }The table separates VLM-facing prompts from deterministic post-generation projection.}
\label{tab:cot-prompt-templates}

{
\footnotesize

\begin{tabularx}{\textwidth}{
    @{}
    >{\raggedright\arraybackslash\bfseries}p{0.13\textwidth}
    >{\raggedright\arraybackslash}X
    >{\raggedright\arraybackslash}X
    @{}
}
\toprule
Prompt component
& LIBERO: single-arm annotation
& RoboTwin: bimanual annotation \\
\midrule

Role
& First infer the global subtask flow of a single-arm episode from selected segment images. Then refine each candidate segment using its main-view and wrist-view evidence together with rule-derived robot context.
& Adjudicate one candidate segment of a bimanual trajectory from synchronized temporal context. Treat robot state/action as the physical anchor and the three camera views as semantic evidence. \\

Visual organization
& Episode planning receives paired images in the order main camera then wrist camera for every selected segment. Segment refinement receives one main-view keyframe and one wrist-view keyframe for the current phase segment.
& The input is a contact sheet. Each row is one draft segment; columns are main, left wrist, and right wrist. Each cell contains synchronized context frames sampled from segment start through end. \\

Context evidence
& Full instruction; rule-derived segment hints from robot state, gripper state, and end-effector motion; ordered task-plan draft; candidate phase, target, location, goal, and contact state.
& Instruction; episode draft summary; candidate frame range; previous adjudicated segments; raw-parquet gripper start/end/trend and closed fraction; left/right arm motion; state/action semantics. \\

Evidence use
& Treat rule hints and the ordered task plan as weak anchors. Use the full instruction and images to correct the global object order, target, destination, phase, and contact interpretation.
& Treat parquet-derived gripper transitions and dominant arm motion as hard physical anchors. Use the main view for object identity, destination, and progress, and retrieve left- and right-wrist evidence separately. \\

Temporal reasoning
& Compare the beginning, middle, and end of the segment with its neighbors. Preserve physically ordered transitions such as approach $\rightarrow$ grasp $\rightarrow$ carry $\rightarrow$ release, and do not create a new subtask for short retry or jitter.
& Reason over synchronized context frames and preceding segments. Preserve handover and control sequences, distinguish simultaneous from inactive arms, and avoid turning brief adjustment or post-release jitter into an independent task-level stage. \\

VLM output schema
& Episode planning returns \texttt{subtasks} and segment objects containing \texttt{segment\_id}, frame range, phase, target/location, goal/location, contact, next trend, reasoning, and confidence. Segment refinement separately returns one JSON object containing target/location, goal/location, phase, subtask, main/wrist observations, \texttt{wrist\_focus}, contact, next trend, and confidence.
& Return \texttt{segments} with \texttt{segment\_id}, frame range, phase, \texttt{active\_arm}, subtask, reasoning, dual-view wrist text, target/location, goal/location, contact, next trend, confidence, state/main/left-wrist/right-wrist evidence, and risk flags; optionally return episode flags. \\

Generation constraints
& Before gripper closure, contact must remain \texttt{not\_in\_contact} or \texttt{touching}; approach cannot claim goal placement. Keep the manipulated object distinct from its location and destination, and keep reasoning robot/gripper-centric and concise.
& An open gripper cannot grasp, hold, lift, or carry. Closing transitions indicate contact/grasp stabilization before transport. Opening after a prior hold indicates release. Dominant-arm motion constrains \texttt{active\_arm}, and the receiving gripper cannot claim a handover grasp before contact. \\

Wrist language
& Return compact \texttt{wrist\_focus} describing wrist-local evidence, such as target proximity, fingertip contact, grasp stability, alignment, finger opening, gripper--object separation, or push/pull motion.
& Always return \texttt{left=...; right=...}, with left before right. Each side describes only its own local evidence and motion; a physically inactive side is written as \texttt{keeps still}. \\

Post-VLM projection
& Validate JSON against rule segments, normalize fields, rebuild compact robot-centric reasoning, repair physically inconsistent phase/target transitions, smooth non-semantic retry or jitter, and expand accepted segment labels to frames.
& Validate ranges and evidence fields, apply per-arm state guards, release-history checks, handover/control projections, temporal smoothing, wrist locality, and frame expansion. A separate normalizer enforces consistent approach, grasp, carry, hold, release, move-away, and stillness wording. \\

Final rendering
& \multicolumn{2}{
    >{\raggedright\arraybackslash}p{0.81\textwidth}@{}
}{
    After projection and normalization, retained labels are rendered as one ordered training string: \texttt{Subtask: ... Reasoning: ... Wrist: ...}. Internal fields such as phase, target, contact state, confidence, and evidence remain metadata rather than train-text fields.
} \\
\bottomrule
\end{tabularx}
}
\end{table}

\begin{figure}[!t]
\centering
\includegraphics[width=1\textwidth]{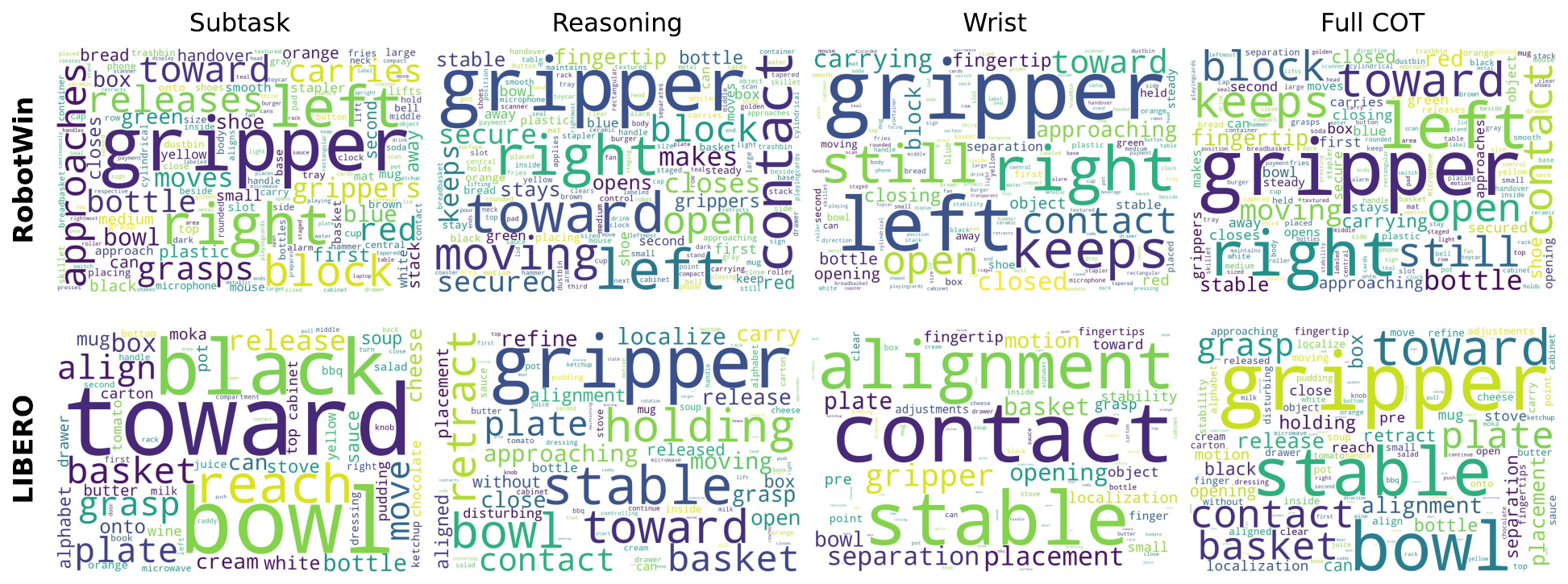}
\caption{\textbf{Word-cloud visualization of the \CoT annotations.}
Rows correspond to datasets, and columns correspond to \textit{Subtask}, \textit{Reasoning}, \textit{Wrist}, and Full CoT.
Word size is proportional to token frequency.}
\label{fig:wordclouds}
\end{figure}

Fig.~\ref{fig:wordclouds} summarizes the vocabulary distribution across the three annotation fields for LIBERO and RoboTwin, highlighting our dense and physically grounded manipulation supervision.
Rather than describing only static objects, the annotations cover a broad set of actionable arm and gripper behaviors, including \textit{reach/approach}, \textit{grasp/pick}, \textit{move/carry}, \textit{align/place}, \textit{release/retract}, \textit{open/close}, \textit{contact}, \textit{secure}, \textit{hold}, and \textit{handover}.
The \textit{Subtask} field captures high-level manipulation progress, the \textit{Reasoning} field encodes state transitions and contact dynamics, and the \textit{Wrist} field grounds the language in local visual evidence such as \textit{contact}, \textit{gripper} \textit{opening/closing}, \textit{object localization}, \textit{alignment}, and \textit{placement stability}.
Overall, \CoT provides structured auxiliary supervision for task decomposition and subtask-level reasoning.
The supervision captures manipulation progress, physical state transitions, and wrist-local evidence throughout each trajectory.

\section{Additional Implementation Details}
\label{app:implementation-details}

\subsection{Model Architecture.}
We instantiate \ours with Qwen3-VL-4B-Instruct as the vision-language backbone and a DiT-B flow-matching action head.
The Qwen3-VL backbone has approximately 4.44B parameters.
Its language model contains 36 transformer layers with hidden dimension 2560, 32 attention heads, and 8 key-value heads.
Its vision encoder contains 24 transformer layers with hidden dimension 1024, patch size 16, temporal patch size 2, and projects visual features into the 2560-dimensional language hidden space.
All Qwen3-VL forward passes are run in bfloat16.

The action head is a DiT-B flow-matching module with 16 transformer blocks.
The action transformer uses a 768-dimensional internal token width, 12 attention heads, adaptive timestep normalization, dropout 0.2, and interleaved self-attention/cross-attention blocks.
For LIBERO, it predicts 7-dimensional delta end-effector actions: 3 Cartesian translation dimensions, 3 axis-angle rotation dimensions, and 1 gripper dimension. For RoboTwin 2.0 and the real-world tasks, it predicts 14-dimensional absolute joint-position actions, with 7 dimensions for each arm including its gripper.
The action head also uses 32 learned future/action query tokens before the action tokens.
The wrist branch uses a frozen V-JEPA 2.1 encoder, a four-layer predictor, and a wrist-context adapter that converts the predicted future wrist latents into 32 future-aware context tokens.

\subsection{Information-Flow Masks for \texorpdfstring{\ours}{W2-VLA} Training and Inference}
\begin{figure}[!t]
    \centering
    \includegraphics[width=0.75\linewidth]{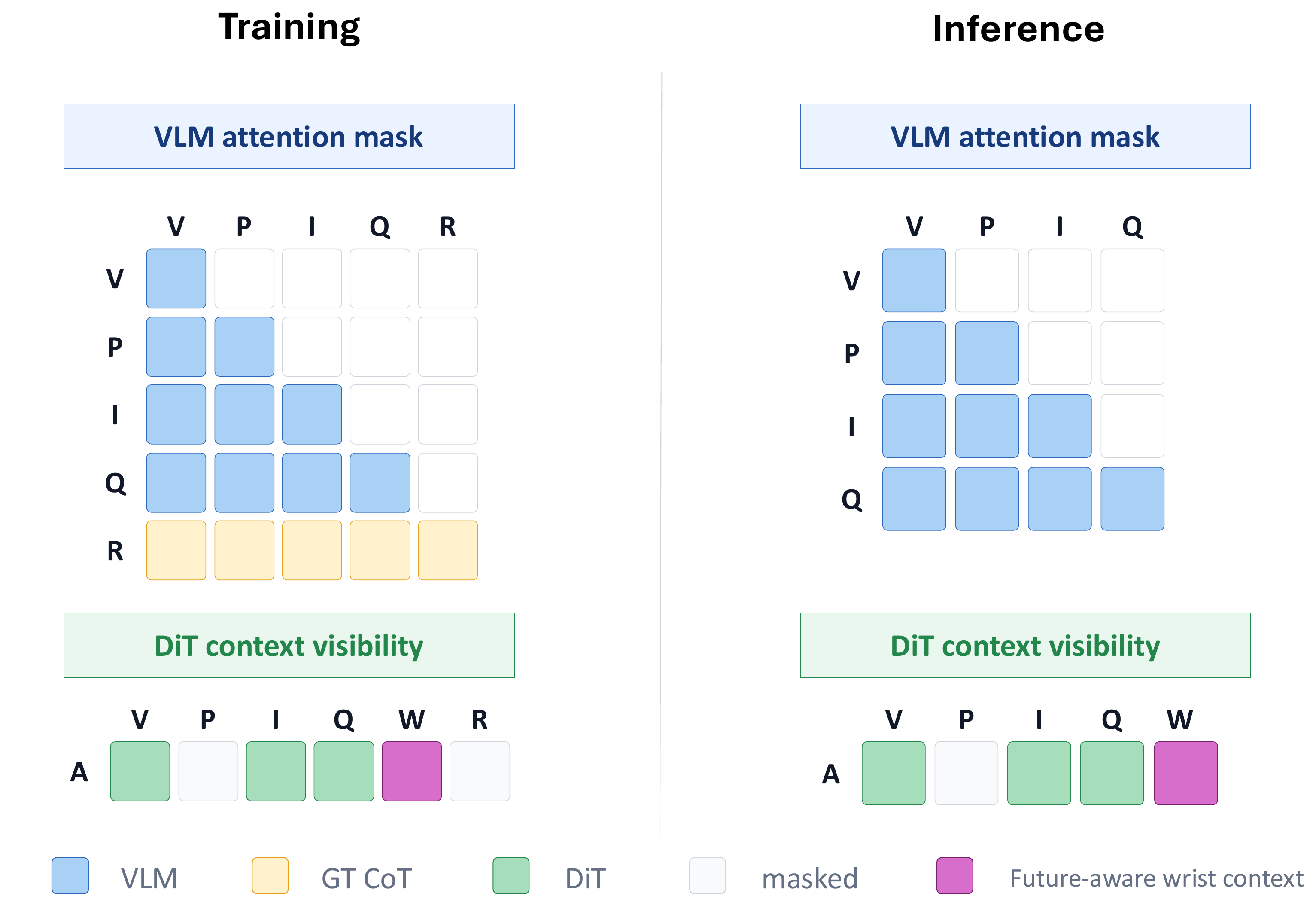}
    \caption{\textbf{Training and inference information-flow masks.}
    During training, structured annotation targets provide auxiliary next-token
    prediction supervision, while prompt-formatting and annotation-target states
    are excluded from the DiT action context.
    The action head is conditioned on visual, instruction, and latent modeling
    states together with the future-aware wrist context produced by the adapter.
    At inference, no annotation sequence is decoded, and the same deployable
    action context is used.}
    \label{fig:information_flow_masks}
\end{figure}

Fig.~\ref{fig:information_flow_masks} illustrates the information-flow masks used during training and inference.
We denote visual tokens by $V$, the CoT prompt template by $P$, the task instruction by $I$, latent modeling tokens by $Q$, structured annotation target tokens by $R$, the predicted future wrist latent by $\hat{\bm{Y}}_t^w$, its adapter-produced future-aware context by $W$, and the Diffusion Transformer action head by $A$.
During training, the VLM is optimized with an auxiliary next-token prediction objective over the structured annotation targets $R$, conditioned on the current visual context, prompt template, task instruction, and latent modeling tokens.
The DiT action head does not receive hidden states from $P$ or $R$.
Its context contains only the deployable tokens from $V$, $I$, and $Q$, together with the future-aware wrist context $W$ produced by the wrist-context adapter from $\hat{\bm{Y}}_t^w$.
This separation keeps structured annotations as training-only supervision, while maintaining consistent state selection and action-context composition between training and inference.

At inference time, no annotation sequence is decoded and the VLM processes $[V, P, I, Q]$.
The prompt template still structures the latent modeling interface, but its hidden states are excluded from the DiT context.
The wrist predictor produces $\hat{\bm{Y}}_t^w$, which the wrist-context adapter converts into $W$ before it is appended to the action context.
Therefore, the DiT action head is conditioned on the same deployable context as in training, namely $V$, $I$, $Q$, and $W$.
This separation uses explicit CoT text only as training supervision, while the deployed policy acts from the latent modeling interface and adapted future-wrist context.

\subsection{Training Details and Simulation Setup}
The complete objective is
\[
\mathcal{L}
= \mathcal{L}_{\mathrm{act}}
+ \lambda_{\mathrm{cot}}\mathcal{L}_{\mathrm{cot}}
+ \lambda_{\mathrm{wrist}}\mathcal{L}_{\mathrm{wrist}},
\]
where $\lambda_{\mathrm{cot}}=0.1$ and $\lambda_{\mathrm{wrist}}=0.2$.
The Qwen backbone, JEPA predictor, wrist-context adapter, and action head are trainable, while V-JEPA 2.1 is frozen.
We train with AdamW using $\beta_1=0.9$, $\beta_2=0.95$, $\epsilon=10^{-8}$, weight decay $10^{-8}$, and gradient clipping at 1.0.
We use a cosine learning-rate schedule with a minimum learning rate $10^{-6}$ and 5000 warmup steps.
The learning rates are $1.0\times 10^{-5}$ for Qwen3-VL, $5.0\times 10^{-5}$ for the JEPA predictor, $5.0\times 10^{-5}$ for the wrist-context adapter, and $1.0\times 10^{-4}$ for the action head.
Both the main view and the wrist view are resized to $224 \times 224$.

At the beginning and end of each episode, the number of available historical wrist views or future views may be shorter than the prediction horizon.
In such cases, we pad the sequence with the available boundary images to match the required length for training.
We further apply a discount coefficient to the JEPA prediction loss to control the learning strength.

For LIBERO, we train for 60K optimization steps on 4 A100 GPUs with a per-device batch size of 16.
The prediction horizon is set to $h=8$, where the model takes 8 historical wrist-view frames as input and predicts 8 future frames.

For RoboTwin 2.0, we train for 100K optimization steps on 8 B200 GPUs with a per-device batch size of 16.
The prediction horizon is set to $h=16$; however, both historical and future wrist frames are sampled every other frame, keeping the actual number of input and predicted frames at 8.

\section{Real-world Experimental Details}
\label{app:real-world-details}

\subsection{Experimental Setup}
\begin{figure}[!t]
\centering
\includegraphics[width=1\textwidth]{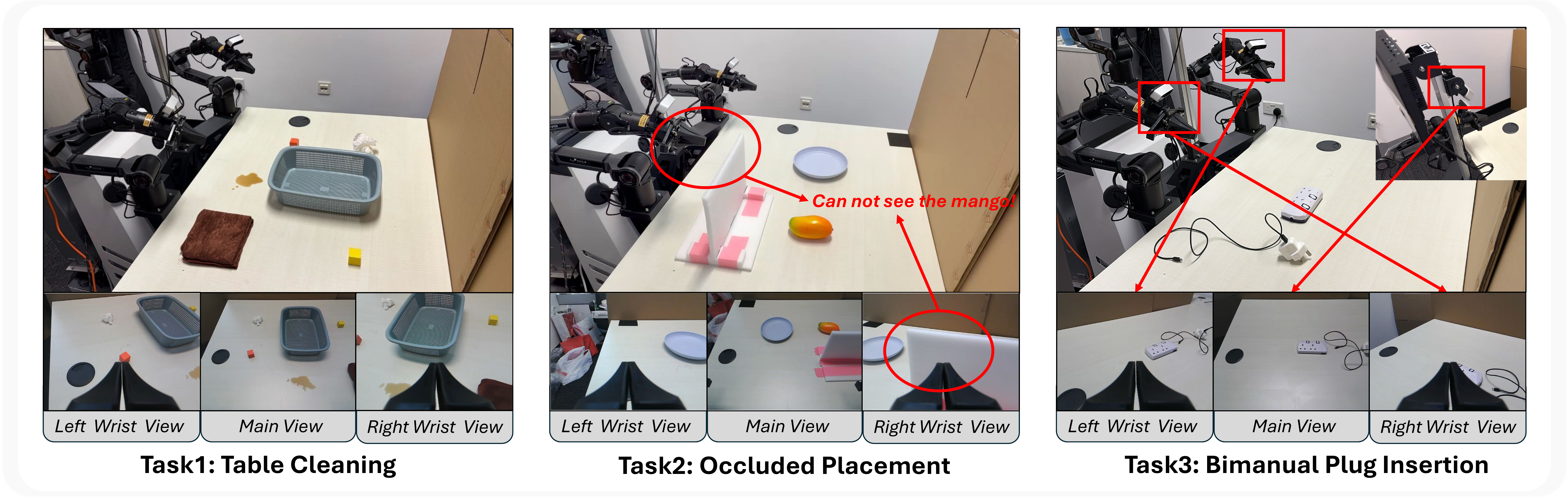}
\caption{\textbf{Real-world Experiments Setup.}
}
\label{fig:real_setup}
\end{figure}

In this subsection, we introduce the three real-world tasks and their experimental settings.
Fig.~\ref{fig:real_setup} shows the experimental setup for deploying the three real-world tasks.

Specifically, \textit{1) Table Cleaning} requires long-horizon stage decomposition, sequential object handling, and switching between the two arms;
\textit{2) Occluded Placement} directly tests complementary global and local visual grounding: the wrist cameras initially do not observe the object, so the main view provides global localization and route context, while the wrist views provide local evidence for grasping, alignment, and release;
\textit{3) Bimanual Plug Insertion} is a fine-grained, contact-rich manipulation task that demands tightly coupled bimanual stabilization and precise 6-DoF plug control, as even small translational or angular errors can cause misalignment, slippage, or jamming.
The main view resolves the global layout and task stage, while the wrist views provide close-range cues for subtle pose corrections, socket-level alignment, contact transitions, and insertion-depth control.

\noindent\textbf{$\bullet$ Table Cleaning.}
The workspace contains a paper ball, two blocks, a small basket, a cloth, and a coke spill.
The left arm sequentially places the paper ball and one of the blocks into the basket.
The right arm then places another block into the basket, then picks up the cloth, and wipes the spill directly in front of the robot until it is removed or substantially covered.
The task succeeds only if all three objects are in the basket and the spill is removed or sufficiently covered.
The main view identifies the multiple objects, basket, and spill region, while the wrist views verify grasp stability, object placement in the basket, cloth--table contact, and coverage of the wiping trajectory.
This task evaluates long-horizon stage decomposition, bimanual task switching, and ordered execution across objects.

The instructions for the table-cleaning task are as follows: \textit{``pick up the crumpled paper and small blocks from the tabletop, place them into the tray, then use the cloth to wipe the brown stain on the table''}

\noindent\textbf{$\bullet$ Occluded Placement.}
We place a foam mango, a plastic plate, a low foam obstacle, and table-positioning tape in the scene.
As shown in Fig.~\ref{fig:real_setup}, initially, neither wrist view observes the mango or the plate, whereas the main view observes the global layout.
The main view provides global context for localizing the mango and plate and for choosing a route that goes around or over the obstacle.
As the robot approaches the mango and plate, the wrist views provide local evidence for grasping, obstacle clearance, alignment, and release.
The task succeeds if the mango is placed on the plate without knocking over the obstacle, dropping the mango from the table, or allowing the mango to roll off the plate.
This task evaluates the task-conditioned World-to-Wrist pathway: the main view provides global target and route context, while the wrist views provide local interaction evidence.

The instructions for the occluded-placement task are as follows: \textit{``pick up the mango and place it into the plate without knocking over the obstacle''}

\noindent\textbf{$\bullet$ Bimanual Plug Insertion.}
We use an unpowered power strip and a corded plug.
The left arm grasps the power strip and holds it fixed throughout the task.
The right arm reaches for and grasps the plug, moves it above the target socket, aligns it with the socket, and then inserts it.
The task succeeds if the left arm keeps the power strip stable while the right arm inserts the plug into the target socket.
This task highlights bimanual coordination and wrist-level, contact-rich alignment: the main view identifies the plug, power strip, and task stage, while the wrist views support local alignment, pre-contact adjustment, and insertion-depth control.

The instructions for the bimanual plug-insertion task are as follows: \textit{``Hold the power strip steady with the left arm, then use the right arm to reach for and grasp the plug, move it above and align it with the target socket, and insert it''}

We additionally design experiments under three OOD settings:
\textit{1) Table Clutter}: randomly placing irrelevant objects on the table;
\textit{2) Random Lighting Perturbations}: using a rotating colored light to
perturb the camera observations; and
\textit{3) Background Variations}: changing the color of the tablecloth.

\section{Further Analysis and Visualizations}
\label{appendix:c}
\subsection{How effective is the wrist predictor?}
\begin{figure}[!htbp]
\centering
\includegraphics[width=0.96\linewidth]{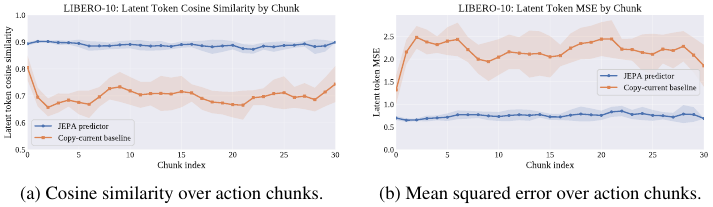}
\caption{\textbf{Effectiveness of the wrist predictor. }Chunk-level latent prediction quality measured by cosine similarity and mean squared error.}
\label{fig:wrist-predicto}
\end{figure}

As shown in Fig.~\ref{fig:wrist-predicto}, we evaluate the quality of the JEPA wrist predictor on LIBERO-10 by comparing the predicted future wrist latent tokens with the ground-truth future wrist latent tokens.
The ground-truth tokens are obtained by encoding the observed future wrist frames with the same visual encoder.
As a simple baseline, we copy the current wrist latent tokens forward and compare them with the same future ground-truth tokens.
We report latent-token mean squared error (MSE) and cosine similarity.

A lower MSE and a higher cosine similarity indicate stronger consistency between the predicted future wrist latent tokens and the wrist observations from the actual rollout.
Across 500 episode records, the JEPA predictor achieves an average latent-token MSE of 0.749, while the copy-current baseline obtains 2.183.
The cosine similarity also improves from 0.699 for the baseline to 0.888 for JEPA, giving a gain of 0.189.
These results show that the JEPA wrist predictor is effective: it predicts future wrist latent tokens that are much closer to the ground-truth future tokens than simply copying the current representation.
The consistently lower MSE and higher cosine similarity indicate that the learned predictor captures useful future visual dynamics in the wrist-view latent space.

\subsection{More visualizations of latent-modeling-to-image attention.}

\begin{figure}[!htbp]
\centering
\includegraphics[width=\linewidth]{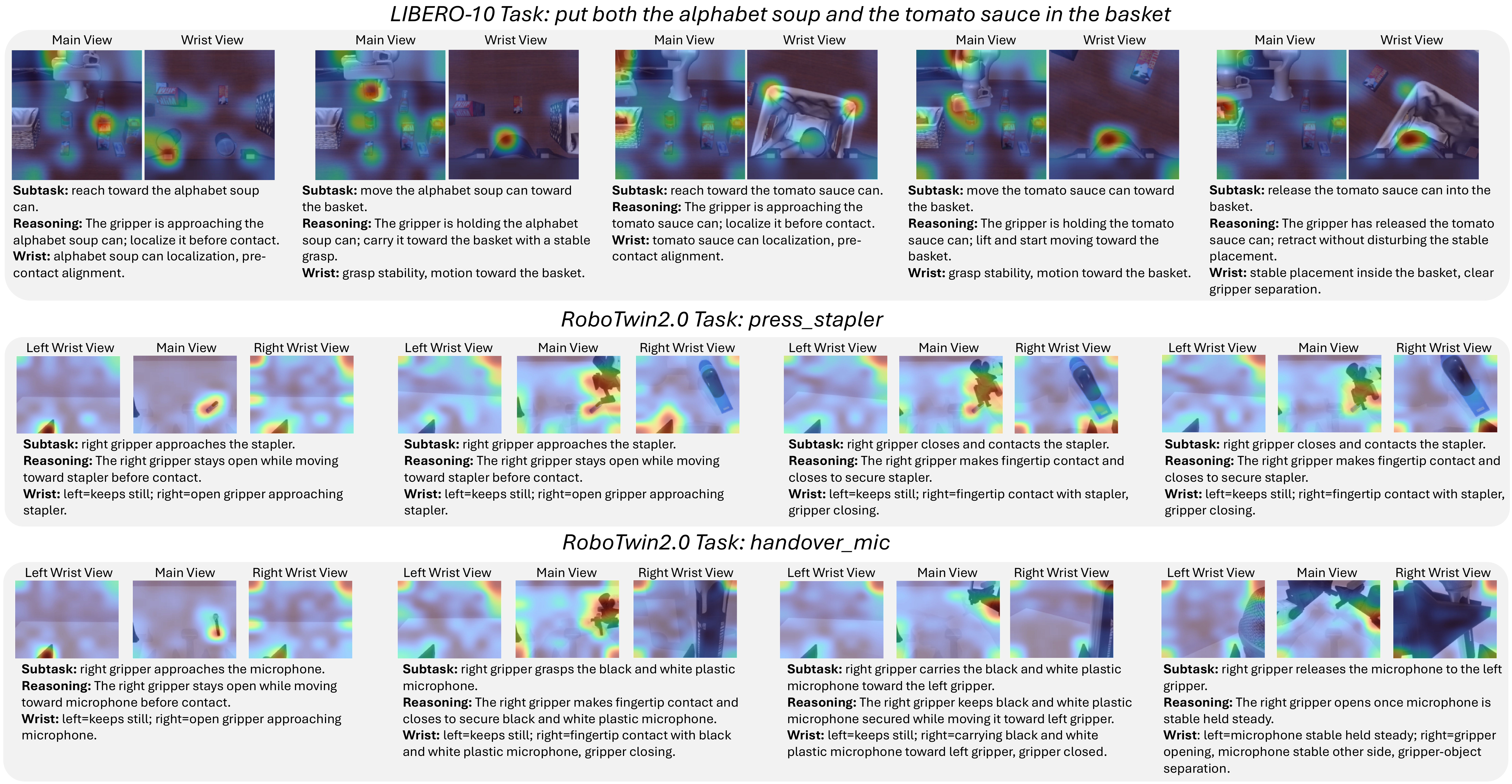}
\caption{\textbf{Attention weights from latent modeling tokens to image tokens.}
We visualize how latent modeling tokens attend to main- and wrist-view image tokens over three task rollouts, and pair the maps with \texttt{Subtask}/\texttt{Reasoning}/\texttt{Wrist} descriptions decoded from the full VLM context for semantic interpretation.
The maps use direct final-layer post-softmax weights averaged over all 32 heads and 16 latent modeling tokens; no gradients or attention rollout are used.
Brighter regions indicate higher relative attention within each independently normalized panel.}
\label{fig:attention_visualizations}
\end{figure}

Fig.~\ref{fig:attention_visualizations} visualizes the attention weights assigned by the latent modeling tokens to image tokens during policy rollouts.
We show how the latent modeling tokens attend to image tokens from the main and wrist views for one LIBERO-10 task and two RoboTwin 2.0 tasks.
For semantic interpretation, we pair the attention maps with the corresponding \texttt{Subtask}/\texttt{Reasoning}/\texttt{Wrist} descriptions generated from the full VLM context, which provide semantic references for the current manipulation stage.

Specifically, we extract direct post-softmax self-attention weights from the final (36th) language-transformer layer of Qwen3-VL (zero-based layer index 35), rather than from the vision encoder or the DiT action head.
For each image token, we uniformly average the attention weights over all 32 attention heads and all 16 learned latent modeling tokens; no individual head or latent token is selected.
The weights are obtained with \texttt{output\_attentions=True} in a no-gradient forward pass using eager attention.
They are single-layer attention probabilities, not gradients, gradient-weighted maps, or attention rollout across layers.

The current RGB views are resized to $224\times224$ and tokenized into an $8\times8$ image-token grid after Qwen's spatial merging.
LIBERO uses the main and wrist views, whereas RoboTwin uses the main, left-wrist, and right-wrist views.
We reshape the averaged scores to the corresponding grid, independently min-max normalize each map, bicubically upsample it, and overlay it on the input image with opacity 0.5.
Therefore, brightness indicates relative spatial attention within one panel and should not be used to compare absolute attention magnitude across views or rollout steps.
Each panel corresponds to one policy call without temporal averaging.
The accompanying CoT description is decoded separately for interpretation and is not used to compute the attention map.

Across the three tasks, the image-token attention assigned by the latent modeling tokens is concentrated on regions relevant to the current stage, including target objects, active grippers, and local contact areas.
In the LIBERO-10 task, the highlighted regions shift between the active can, the gripper, and the basket as the policy progresses from approach and grasp to transport and release.
For \texttt{press\_stapler}, attention emphasizes the approaching gripper and the stapler before becoming concentrated around their contact region.
For \texttt{handover\_mic}, it follows the microphone and the two grippers across approach, grasp, transfer, and release.

\subsection{Efficiency Analysis}
\begin{figure}[H]
\centering
\includegraphics[width=0.58\linewidth]{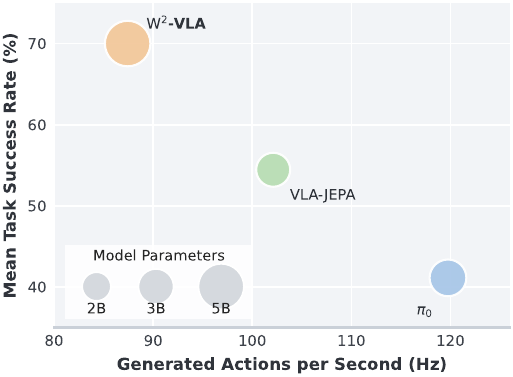}
\caption{\textbf{Efficiency comparison.}
Mean standard-condition success rate is plotted against generated actions per second, while bubble area represents the total number of model parameters.
The generated-actions-per-second metric is computed as the action-chunk length divided by the latency for generating one complete chunk.}
\label{fig:efficiency_analysis}
\end{figure}

Fig.~\ref{fig:efficiency_analysis} compares mean standard-condition success, action throughput, and model size in the real-world experiments.
We measure throughput by amortizing action-chunk generation latency over all actions in the generated chunk.
Specifically, if a policy generates a chunk of $L$ actions in $T_{\mathrm{chunk}}$ seconds, we compute its action-generation throughput as
\begin{equation}
    R_{\mathrm{act}} = \frac{L}{T_{\mathrm{chunk}}}.
\end{equation}
This metric quantifies action-generation capacity rather than the number of policy forward passes per second.
Using the measured chunk-generation latencies, \ours generates a 16-action chunk in $183\,\mathrm{ms}$, yielding $16/0.183=87.43\,\mathrm{Hz}$.
$\pi_0$ generates a 50-action chunk in $417.55\,\mathrm{ms}$, yielding $50/0.41755=119.75\,\mathrm{Hz}$, while VLA-JEPA generates a 7-action chunk in $68.55\,\mathrm{ms}$, yielding $7/0.06855=102.12\,\mathrm{Hz}$.
Thus, although \ours has lower peak throughput, all three methods operate in a comparable high-frequency regime ($87.43$--$119.75\,\mathrm{Hz}$), and \ours retains sufficient action-generation capacity to support real-time control in our deployment.
More importantly, the higher generation throughput of the baselines does not translate into higher real-world task completion: \ours achieves a mean standard-condition success rate of $70.00\%$, outperforming VLA-JEPA ($54.44\%$) by $15.56$ percentage points and $\pi_0$ ($42.22\%$) by $27.78$ percentage points.

\clearpage
\subsection{Visualization of rollouts in real-world OOD scenarios.}
\begin{figure}[H]
\centering
\includegraphics[width=1\linewidth]{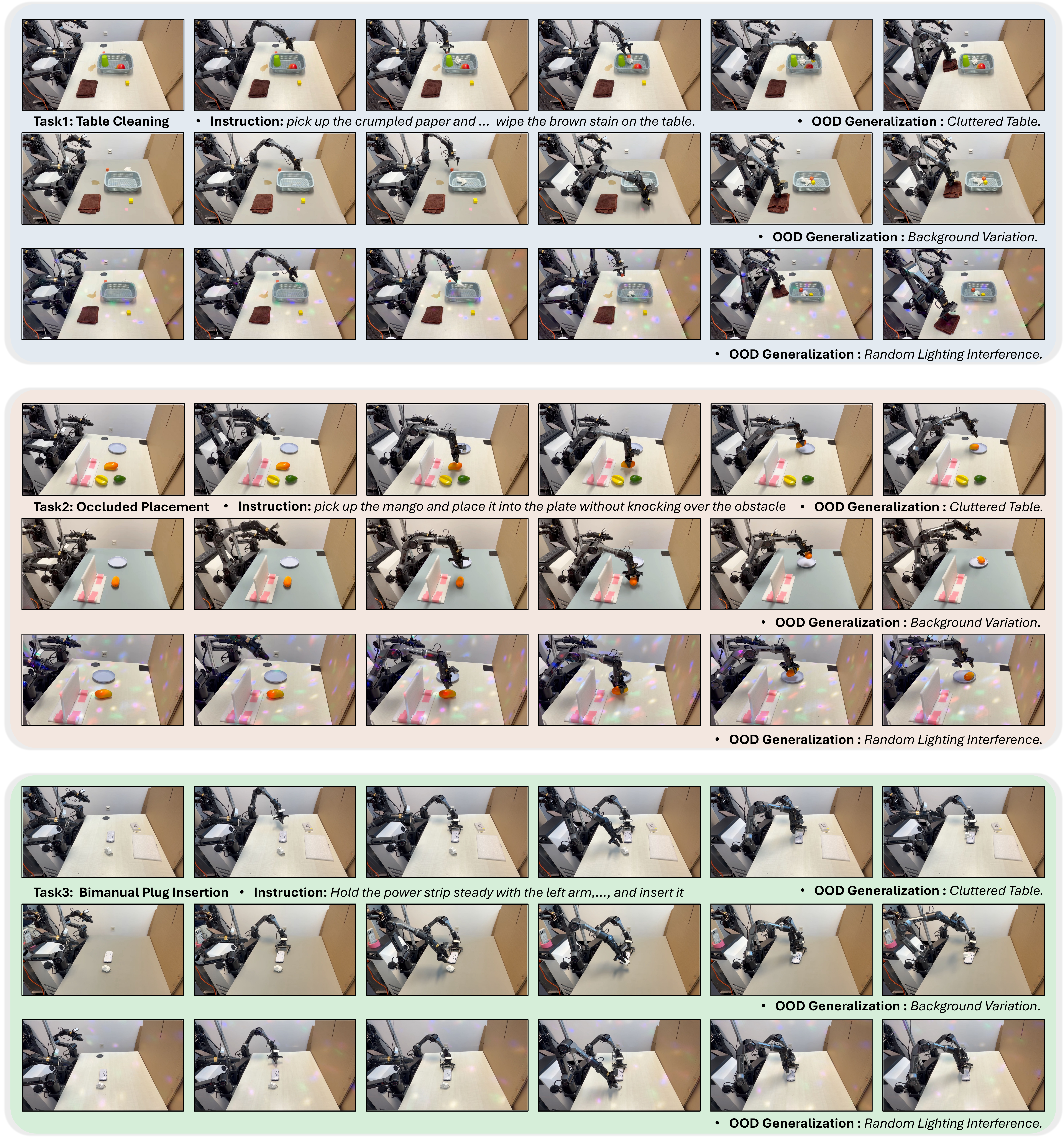}
\caption{\textbf{Rollout Examples in Real-World Tasks.}
We show three OOD settings: 1) Table Clutter, 2) Random Lighting Perturbations, and 3) Background Variations.
}
\label{fig:real_generalization}
\end{figure}

Fig.~\ref{fig:real_generalization} shows rollout image sequences of \ours for the three real-world tasks in the generalization settings.
Our method maintains strong generalization performance under table clutter, lighting perturbations, and background variations.

\FloatBarrier

\end{document}